\documentclass[Afour,sageh,times]{sagej}

\usepackage{moreverb,url}

\usepackage[colorlinks,bookmarksopen,bookmarksnumbered,citecolor=red,urlcolor=red]{hyperref}
\usepackage[linesnumbered,ruled,vlined]{algorithm2e}
\usepackage{amsmath}
\usepackage{tabstackengine}
\usepackage{arydshln}
\usepackage{multirow}
\usepackage{subcaption}
\usepackage{etoolbox}
\usepackage{mathtools}

\newcommand\BibTeX{{\rmfamily B\kern-.05em \textsc{i\kern-.025em b}\kern-.08em
T\kern-.1667em\lower.7ex\hbox{E}\kern-.125emX}}

\begin{document}

\runninghead{Anqing Duan}

\title{A Structured Prediction Approach for Robot Imitation Learning}

\author{Anqing Duan\affilnum{1}, Iason Batzianoulis\affilnum{2}, Raffaello Camoriano\affilnum{3}, Lorenzo Rosasco\affilnum{4, 5, 6},\\ Daniele Pucci\affilnum{7}, and Aude Billard\affilnum{2}}

\affiliation{\affilnum{1}Robotics and Machine Intelligence Laboratory, The Hong Kong Polytechnic University, Hong Kong SAR, China\\ \affilnum{2}Learning Algorithms and Systems Laboratory, École Polytechnique Fédérale de Lausanne, Lausanne, Switzerland\\ \affilnum{3}Visual And Multimodal Applied Learning Laboratory, Politecnico di Torino, Turin, Italy\\ \affilnum{4}DIBRIS, Università degli Studi di Genova, Genoa, Italy\\ \affilnum{5}Laboratory for Computational and Statistical Learning (IIT@MIT), Istituto Italiano di Tecnologia and Massachusetts Institute of Technology, Cambridge, MA, United States\\ \affilnum{6}Machine Learning Genoa (MaLGa) Center, Università di Genova, Genoa, Italy\\ \affilnum{7}Dynamic Interaction Control research line, Italian Institute of Technology, Genoa, Italy}

\corrauth{Raffaello Camoriano, Politecnico di Torino, C.so Francesco Ferrucci 112, 10141 Turin, Italy.}

\email{raffaello.camoriano@polito.it}

\begin{abstract}
We propose a structured prediction approach for robot imitation learning from demonstrations.
Among various tools for robot imitation learning, supervised learning has been observed to have a prominent role.   
Structured prediction is a form of supervised learning that enables learning models to operate on output spaces with complex structures.
Through the lens of structured prediction, we show how robots can learn to imitate trajectories belonging to not only Euclidean spaces but also Riemannian manifolds.
Exploiting ideas from information theory, we propose a class of loss functions based on the $f$-divergence to measure the information loss between the demonstrated and reproduced probabilistic trajectories.
Different types of $f$-divergence will result in different policies, which we call \textit{imitation modes}.
Furthermore, our approach enables the incorporation of spatial and temporal trajectory modulation, which is necessary for robots to be adaptive to the change in working conditions.
We benchmark our algorithm against state-of-the-art methods in terms of trajectory reproduction and adaptation.
The quantitative evaluation shows that our approach outperforms other algorithms regarding both accuracy and efficiency. 
We also report real-world experimental results on learning manifold trajectories in a polishing task with a KUKA LWR robot arm, illustrating the effectiveness of our algorithmic framework.
\end{abstract}

\keywords{Imitation learning, structured prediction, learning and adaptive systems, kernel methods, Riemannian manifolds}
\maketitle

\section{Introduction}\label{intro}
The general notion of imitation is widely exploited in robotics as it can bring multiple benefits \citep{osa2018algorithmic}. 
For example, imitation learning has been proven to be an effective approach to facilitate the acquisition of motor skills for complex high-dimensional humanoid robots \citep{schaal1999imitation, yang2018dmps}.   
Also, imitation learning can be employed for robots to achieve tasks whose rewards are intricate to manually specify, such as aerobatic maneuvers for helicopter flight \citep{abbeel2010autonomous} and dynamic flips and spins \citep{peng2018deepmimic}.
Besides, by improving policies initialized by imitation learning, reinforcement learning can converge faster than optimizing a policy from scratch \citep{kober2014policy, cheng2018fast}.

\begin{table*}[t]
	
	\caption{Comparison of features with respect to state-of-the-art methods.}
	\label{comparison}
	\centering
	\footnotesize
	\begin{tabular}{lccccccc}
		\toprule
		&Probabilistic   &\begin{tabular}{@{}c@{}}Trajectory\\modulation\end{tabular} &\begin{tabular}{@{}c@{}}Manifold \\ output\end{tabular}&\begin{tabular}{@{}c@{}}Manifold \\ modulation\end{tabular} &\begin{tabular}{@{}c@{}}Multiple\\imitation modes\end{tabular}   &\begin{tabular}{@{}c@{}}Multi-dim \\ input\end{tabular} &\begin{tabular}{@{}c@{}}Multiple \\ output types\end{tabular}\\
		\midrule
		DMP \citep{ijspeert2013dynamical}   & -   &\checkmark    &-    &-  &-   &-  &-  \\
		ProMP \citep{paraschos2013probabilistic} & \checkmark   &\checkmark    &-   &-  &-   &- &-\\
		LAT \citep{reiner2014lat}   & \checkmark  &-    &-   &- &-  &- &-\\
		GMM \citep{zeestraten2017approach}   & \checkmark   &-    &\checkmark   &- &-  &\checkmark &-\\
		LGP \citep{schneider2010robot}  & \checkmark   &-   &-    &-  &-  &- &-\\
		TLGC \citep{ahmadzadeh2018trajectory}  & -   &-    &\checkmark   &-  &-  &- &-\\
		KMP \citep{huang2019kernelized}   & \checkmark   &\checkmark     &-   &- &- &\checkmark &-  \\
		LPV-DS \citep{figueroa2018physically} &-    &\checkmark   &\checkmark   &\checkmark  &-  &\checkmark &- \\
		\textbf{Our Approach}  & \checkmark    &\checkmark   &\checkmark   &\checkmark  &\checkmark  &\checkmark &\checkmark  \\
		\bottomrule
	\end{tabular}
\end{table*}

Briefly, there are two major paradigms for implementing imitation given expert demonstrations. 
The first is centered around the policy, where a policy is directly learned by applying a supervised learning algorithm to find a mapping from input states and context factors to output actions \citep{billard2008robot}. 
The second is centered around the reward, where an unknown reward function is recovered through inverse reinforcement learning or inverse optimal control \citep{abbeel2004apprenticeship, ratliff2009learning}.

In this paper, we focus on policy-centered imitation learning.
Particularly, we consider the scenario where policy representation is instantiated with a trajectory-level abstraction.
In this context, imitation learning is also known as programming by demonstration, where a learner's motion skills are usually acquired by penalizing deviation from the demonstrated trajectory.

Notably, movement primitives remain a central research topic in trajectory imitation with the goal of encoding motor skills from demonstrated trajectories for subsequent usage \citep{ravichandar2020recent}.
To this end, various supervised learning algorithms have been leveraged \citep{stulp2015many}.
More specifically, regression techniques, either parametric or non-parametric, constitute a significant contribution to the development of movement primitives. 
In the following, we briefly cover relevant works highlighting their strengths and limitations.

Dynamic Movement Primitives (DMP) is one of the pioneering imitation learning algorithms to mimic the expert’s trajectory \citep{ijspeert2013dynamical}.
It has been gaining popularity, as evidenced by its large number of derivatives, such as sequenced DMP \citep{kulvicius2011joining}, generalized DMP \citep{zhou2017task}, constrained DMP \citep{duan2018constrained}, etc. 
Recent advances in DMP such as Neural Dynamic Policies (NDPs) integrate deep learning to handle high-dimensional inputs like visual data \citep{bahl2020neural}.
The conception of DMP is based on the spring-damper dynamic system, whose acceleration profile is fitted with a set of manually defined basis functions to capture the shape of a demonstrated trajectory.
DMP supports goal adaptation, yet it cannot handle via-point constraints and multiple demonstrations.
To overcome the limitations, Probabilistic Movement Primitives (ProMP) was developed to learn a distribution over trajectories \citep{paraschos2013probabilistic}.
Besides, ProMP can execute via-point trajectory adaptation, achieved by Gaussian conditioning.  

In contrast to DMP and ProMP, motion imitation can also be realized from a non-parametric angle.
For example, Kernelized Movement Primitives (KMP) leverages the kernel trick for movement representation \citep{huang2019kernelized}.
Due to the kernel formulation, it is straightforward for KMP to handle multi-dimensional inputs, which could be exploited in human-robot collaboration or task synergy retrieval \citep{zeestraten2017learning}.
Other non-parametric methods built upon Gaussian processes, such as Local Gaussian process regression (LGP) \citep{schneider2010robot} and Gaussian process Models (GPM) \citep{9345363}, can learn the demonstrated trajectory as well. 

Moreover, autonomous dynamical systems are also powerful tools for imitation learning. 
For example, Stable Estimator of Dynamical Systems (SEDS) \citep{khansari2011learning} or Linear Parameter-Varying Dynamical System (LPV-DS)~\citep{figueroa2018physically} drops explicit time dependency and can ensure global stability while being robust to external disturbances.
Dynamical systems can be extended to control forces at the contact level \citep{amanhoud2019dynamical}. 
Learning approaches can be used to model both the motion and force profile \citep{khoramshahiarm} and adapt to a prescribed surface.

Despite the aforementioned advancements, so far only the imitation of trajectories in Euclidean spaces has been investigated, leaving the issue of learning trajectories with manifold constraints relatively under-explored. 
Arguably, many imitation objectives in robotics involve the analysis of geometry-structured training data, such as rotation matrices \citep{traversaro2016identification}, stiffness ellipsoids \citep{ajoudani2018reduced}, etc.
More importantly, it can be safety-critical to avoid breaking manifold-imposed constraints in some applications \citep{ahmadzadeh2018trajectory, duan2022ultrasound}.  
Driven by theoretical questions and practical gains, it is thus crucial to develop imitation learning algorithms that are applicable to Riemannian manifolds.

In this paper, we present a persistent algorithmic framework for probabilistic imitation learning.
The key novelty in our approach is to adopt a structured prediction formulation for robot imitation learning.
Structured prediction enables complex outputs and can deal with trajectories on manifolds. 
Our proposed approach can handle the imitation of trajectories lying in either Euclidean space or a manifold, whilst also preserving the essential functionalities for movement primitives.  
Thanks to the inherent kernel method, our approach admits a non-parametric formalism and can learn trajectories driven by multi-dimensional inputs. 
 
When carrying out probabilistic trajectory imitation, it is necessary to specify a suitable loss function that measures the discrepancy between the demonstrated and reproduced probabilistic trajectories.  
Following \citet{ke2019imitation} and \citet{ghasemipour2019divergence}, we exploit tools from information theory and consider defining loss functions with $f$-divergences.
Noticeably, a number of existing imitation learning algorithms, developed in the context of sequential decision-making or programming by demonstration, can be seen as the minimization of some $f$-divergence.
For example, behavior cloning minimizes the Kullback-Leibler (KL) divergence \citep{pomerleau1989alvinn}, KMP minimizes the reverse KL divergence \citep{huang2019kernelized}, and GAIL minimizes the Jensen-Shannon (JS) divergence \citep{ho2016generative}. 
We adopt these ideas in the structured prediction framework and show that by using different divergences as loss functions, it is possible to obtain different imitation strategies, that we call \textit{imitation modes}. 
Different imitation modes determine different coupling effects between the mean and the covariance of the obtained probabilistic trajectory policy.

A comparison between our approach and state-of-the-art algorithms is shown in Table \ref{comparison}. 
To summarize, our contribution is the development of a structured prediction framework for robot motion imitation that enables:
\begin{enumerate}
	\item[(\romannumeral 1)] Prediction of outputs of a variety of types, including Euclidean and manifold-structured trajectories;
	\item[(\romannumeral 2)] Imitation of expert demonstrations with multiple imitation modes by means of different $f$-divergences;
	\item[(\romannumeral 3)] Modulation of trajectories to adapt the learned motion skills to novel working settings.
\end{enumerate}

The rest of the paper is organized as follows.
In Section \ref{metaalg}, we present the proposed algorithmic framework, casting imitation learning as a structured prediction problem and constructing loss functions based on $f$-divergences.
Also, strategies for trajectory modulation are provided.
Within the proposed framework, we introduce its practical implementation in Section \ref{prac}, where both Euclidean and manifold-valued outputs are considered. 
The experimental results, including a comparison against previous algorithms and real-world tasks, are reported in Section \ref{eval} to show the effectiveness of the proposed approach.
We review related work and discuss the limitations and future work in Section~\ref{discuss}.
Finally, we conclude the paper in Section \ref{concl}.

\section{A Structured Prediction Approach to Probabilistic Imitation Learning}\label{metaalg}
In this section, we first briefly review the background of probabilistic imitation learning and present our problem formulation (Section \ref{rawpro}).
We then present our algorithmic framework that reveals a structured approach to probabilistic imitation learning (Section \ref{regimi}), followed by trajectory modulation strategies (Section \ref{trajmod}).

\subsection{Background and Problem Setting}\label{rawpro}
Probabilistic approaches are very popular in robot imitation learning \citep{billard2008robot}.
Compared to deterministic techniques, they can provide more information (such as variability and correlation) to a robot learner.
To perform probabilistic imitation, a human teacher usually presents multiple demonstrations for a single task.
Assume that the raw data collected from $M$ demonstrations of fixed length $N$ is formatted as $\{\{\mathbf{x}_{n}^m, \mathbf{y}_{n}^m \}_{n=1}^N\}_{m=1}^M$, where $\mathbf{x}_{n}^m \in \mathcal{X}$ is the input and $\mathbf{y}_{n}^m \in \mathcal{Y}$ denotes the output. 

The data types of the input and the output spaces in robot imitation learning can vary in different application scenarios.
For instance, in a common scenario where the robot needs to learn a time-indexed trajectory, we have $\mathcal{X} = \mathbb{R}$. 
In a more complex task such as human-robot collaboration, the robot could be required to react to the human's position.
As a result, the input that the robot takes is a vector rather than a scalar, i.e., $\mathcal{X} = \mathbb{R}^\mathcal{I}$ with $\mathcal{I}>1$ being the dimensionality.
The output space $\mathcal{Y}$ is usually considered to have a Euclidean structure $\mathbb{R}^\mathcal{O}$ with $\mathcal{O}$ being the dimensionality.
Additionally, often there are problems where manifold-type constraints are enforced on the output value, i.e., $\mathcal{Y} = \mathcal{M}$, where $\mathcal{M}$ denotes a manifold.  

\begin{figure}[t]
	\centering       
	\includegraphics[width=0.5\textwidth]{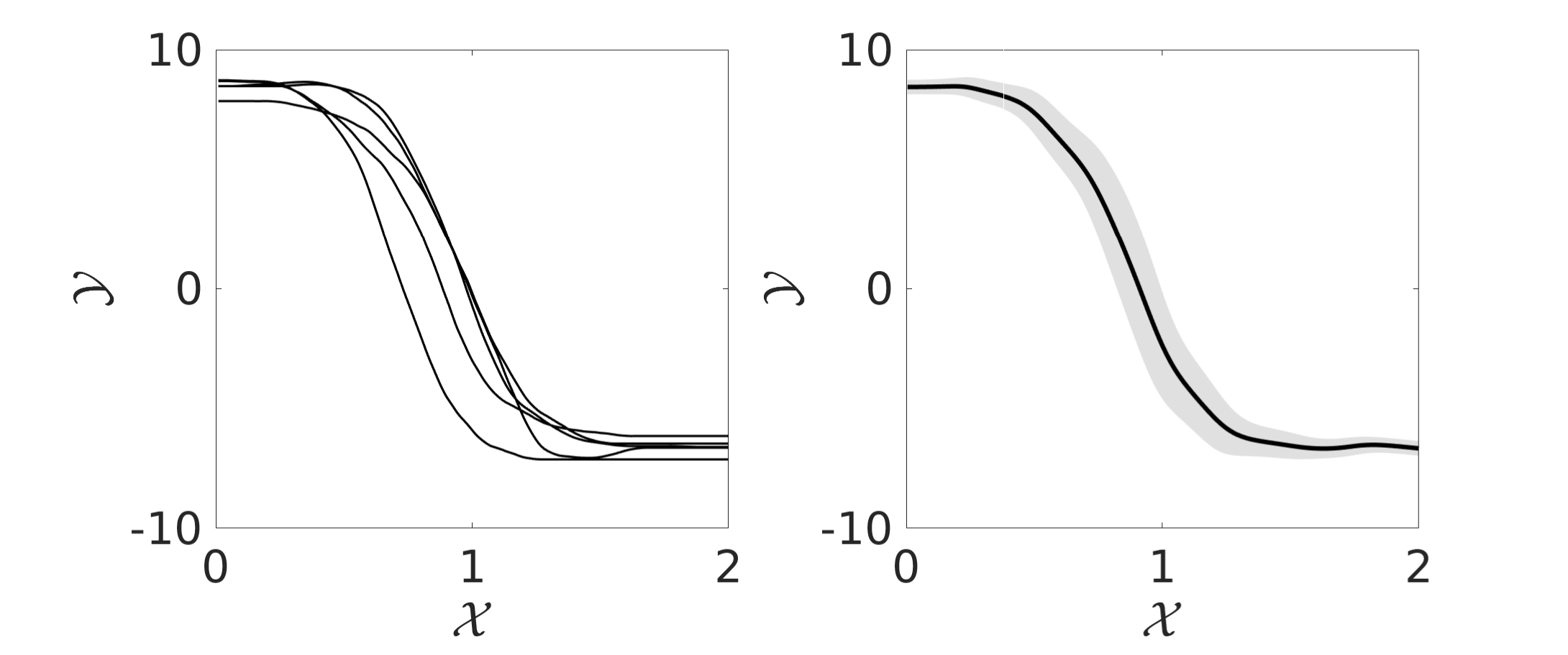}
	\caption{An illustrative example of data processing for probabilistic imitation learning. The \textit{left} figure plots multiple demonstrations and the \textit{right} figure plots the obtained probabilistic trajectory. The solid line represents the mean and the shallow area represents the covariance.}\label{fig:exp_pil}
\end{figure}

To exploit the probabilistic properties from raw demonstration datasets, suitable statistical learning tools such as mixture models \citep{billard2008robot} and their various extensions \citep{zeestraten2017approach, simo20173d} can be employed.  
More precisely, we can have 
\begin{equation}
\{\{\mathbf{x}_{n}^m, \mathbf{y}_{n}^m \}_{n=1}^N\}_{m=1}^M
\xRightarrow[\text{processing}]{\text{data}}
\mathbb{D} = \{\mathbf{x}_n, \tilde{\mathbf{y}}_n\}_{n=1}^N,
\end{equation}
where the output $\tilde{\mathbf{y}}$ in the dataset $\mathbb{D}$ lies in a probability space as a result of multiple demonstrations and can be approximated by some Gaussian-like distribution $\mathcal{P}(\mathcal{Y})$.  
An illustrative example of data processing is shown in Figure \ref{fig:exp_pil}.

In the context of robot movement imitation, a central topic is how to generate a trajectory so that a robot can mimic a demonstrator as closely as possible.
To achieve this goal, the robot should behave as the demonstrator in response to a query input.
 
It is then critical to find a suitable mapping between the input and output values based on the collected demonstrations \citep{zahra2022bio}.
This classical view on imitation learning is reminiscent of the objective of supervised learning: To find an input/output function given input/output pairs. 

Focusing on probabilistic imitation learning, our goal is to address the motion imitation problem by learning the mapping rule $\mathbf{s}: \mathcal{X} \rightarrow \mathcal{P}(\mathcal{Y})$ given the dataset $\mathbb{D}$.
Formally, the problem formulation can be described as 	
\begin{equation}\label{probsetting}
\mathrm{find} \; \mathbf{s}: \mathcal{X} \rightarrow \mathcal{P}(\mathcal{Y}) \quad \mathrm{given} \; \{\mathbf{x}_n, \tilde{\mathbf{y}}_n\}_{n=1}^N,
\end{equation}
where we are looking for a mapping rule with outputs being probability distributions.

\subsection{A Structured Prediction Perspective on Motion Imitation}
\label{regimi}
Before addressing our concerned problem \eqref{probsetting}, we first recall the standard supervised learning setting
\begin{equation}\label{probdet}
	\mathrm{find} \; \mathbf{s}: \mathcal{X} \rightarrow \mathcal{Y} \quad \mathrm{given} \; \{\mathbf{x}_n, \mathbf{y}_n\}_{n=1}^N.
\end{equation}
In particular, following \citep{ciliberto2016consistent}, we call \eqref{probdet} a structured prediction problem whenever the space  $\mathcal{Y}$ does not have a linear structure, e.g., when $\mathcal{Y}$ is manifold. 

\subsubsection{Structured Prediction via Surrogate Approach}
Due to the lack of linearity in the output space, structured prediction is a very challenging problem.
Here, we resort to a surrogate solution that is common in classification \citep{mroueh2012multiclass} and apply it to the structured prediction problem as shown in \citep{ciliberto2016consistent} by leveraging results for vector-valued kernel learning \citep{alvarez2012kernels}.

The key steps of the surrogate approach are sketched as follows:
\begin{enumerate}
	\item \textit{Encoding}.
 Design an encoding  
$\mathbf{c}: \mathcal{Y} \rightarrow \mathcal{H}$
to map  the structured output space  $\mathcal{Y}$ into a vector space $\mathcal{H}$. 
	\item \textit{Surrogate learning}. Solve the learning problem in the surrogate space. This is achieved by first choosing
	a surrogate loss $\mathcal{L}: \mathcal{H} \times \mathcal{H}\rightarrow\mathbb{R}$ and then finding 
 $\mathbf{g}: \mathcal{X}\rightarrow\mathcal{H}$ which minimizes the sum of errors $\mathcal{L}(\mathbf{c}(\mathbf{y}_n), \mathbf{g}(\mathbf{x}_n))$ given the surrogate dataset $\{\mathbf{x}_{n}, \mathbf{c}(\mathbf{y}_{n})\}_{n=1}^N$.
 	\item \textit{Decoding}.  Recover $\mathbf{s}$  with  a suitable decoding rule  $\mathbf{c}^{-1}: \mathcal{H}\rightarrow\mathcal{Y}$, i.e., $\mathbf{s} = \mathbf{c}^{-1}\circ\mathbf{g}:\mathcal{X} \rightarrow \mathcal{Y}$.
	
 \end{enumerate}

\begin{figure}[t]
	\centering       
	\includegraphics[width=0.39\textwidth]{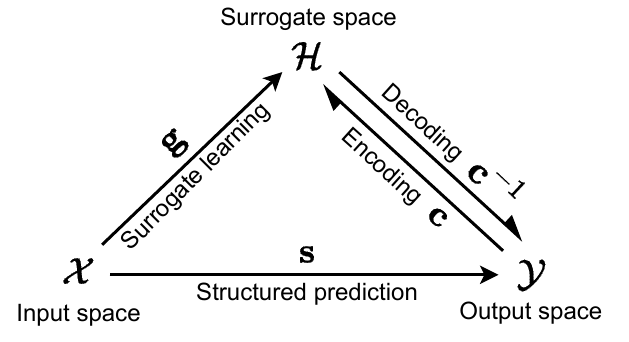}
	\caption{Schematic illustration of the surrogate approach to structured prediction.}\label{fig:surrodiag}
\end{figure}

A pictorial illustration of structured prediction by the surrogate framework is shown in Figure \ref{fig:surrodiag}. 
Next, we discuss how the encoding steps can be performed {\em implicitly} for a wide spectrum of loss functions. 

\subsubsection{Implicit Encoding Framework}\label{ILE}  
As discussed in \citep{ciliberto2020general}, the embedding can be applied implicitly for the case of Structure Encoding Loss Functions (SELF)
$\Delta: \mathcal{Y} \times \mathcal{Y} \rightarrow \mathbb{R}$ for which
 there exists a separable Hilbert space $\mathcal{H}_\mathcal{Y}$ with inner product $\langle \cdot, \cdot \rangle_{\mathcal{H}_\mathcal{Y}}$, a continuous feature map $\mathbf{c}: \mathcal{Y} \rightarrow \mathcal{H}_{\mathcal{Y}}$, and a continuous linear operator $V: \mathcal{H}_{\mathcal{Y}} \rightarrow \mathcal{H}_{\mathcal{Y}}$ such that for all $\mathbf{y}$, $\mathbf{y}' \in \mathcal{Y}$ we have  
\begin{equation}\label{defSELF}
\Delta(\mathbf{y}, \mathbf{y}') = \langle \mathbf{c}(\mathbf{y}), V\mathbf{c}(\mathbf{y}') \rangle_{\mathcal{H}_{\mathcal{Y}}}.
\end{equation}

To address the surrogate learning problem, we consider the following linearly parameterized model 
\begin{equation}\label{gpara}
\mathbf{g}(\mathbf{x}) = \mathbf{W}\boldsymbol{\varphi}(\mathbf{x}) \in \mathbb{R}^{M},
\end{equation}
where $\boldsymbol{\varphi}: \mathcal{X}\rightarrow\mathbb{R}^P$ denotes a feature map and $\mathbf{W}\in \mathbb{R}^{M \times P}$ denotes learnable parameters. 
The estimation of $\mathbf{W}$ can be determined by addressing the following multi-variate ridge regression problem:
\begin{equation}\label{minoptW}
\underset{\mathbf{W}}{\mathtt{\min}} \frac{1}{N}\sum_{n=1}^{N}\|\mathbf{W}\boldsymbol{\varphi}(\mathbf{x}_n) - \mathbf{c}(\mathbf{y}_n)\|_{ \mathcal{H}_{\mathcal{Y}}}^2 + \lambda \|\mathbf{W}\|_{F}^2.
\end{equation}
The solution to \eqref{minoptW} can be shown to be
\begin{equation}\label{gW}
\widehat{\mathbf{W}} = \mathbf{C}(\boldsymbol{\Phi}^\top\boldsymbol{\Phi} + N\lambda \mathbf{I}_N)^{-1}\boldsymbol{\Phi}^\top,    
\end{equation}
where we denote 
$\mathbf{C} = \begin{bmatrix}\mathbf{c}(\mathbf{y}_1), \ldots, \mathbf{c}(\mathbf{y}_N)\end{bmatrix} \in \mathbb{R}^{M \times N}$
and
$\boldsymbol{\Phi} = \begin{bmatrix}\boldsymbol{\varphi}(\mathbf{x}_1), \ldots, \boldsymbol{\varphi}(\mathbf{x}_N)\end{bmatrix} \in \mathbb{R}^{P \times N}$.
Also,
$\lambda > 0$ is a regularization parameter and
$\mathbf{I}_N \in \mathbb{R}^{N \times N}$ denotes the identity matrix of size $N$.
Besides, $\|\cdot\|_F^2$ denotes the squared Frobenius norm of a matrix, i.e., the sum of all its squared elements.

By substituting \eqref{gW} into \eqref{gpara}, the solution to the surrogate learning problem is given by 
\begin{equation}\label{surrofunc}
\widehat{\mathbf{g}}(\mathbf{x}) = \widehat{\mathbf{W}} \boldsymbol{\varphi}(\mathbf{x}) = \sum_{n=1}^{N}\boldsymbol{\alpha}_n(\mathbf{x})\mathbf{c}(\mathbf{y}_n),
\end{equation}
where $\boldsymbol{\alpha}_n(\mathbf{x})$ is the $n$-th entry of $\boldsymbol{\alpha}(\mathbf{x}) \in \mathbb{R}^{N}$: 
\begin{align}\label{beforekernel}
	\boldsymbol{\alpha}(\mathbf{x}) &= (\boldsymbol{\Phi}^\top\boldsymbol{\Phi} + N\lambda \mathbf{I}_N)^{-1}\boldsymbol{\Phi}^\top\boldsymbol{\varphi}(\mathbf{x})  \\
 &= (\mathbf{K}+N\lambda \mathbf{I}_N)^{-1}\mathbf{k}_x, \label{weightsalphacal}   
\end{align}
where the kernel trick is invoked to obtain \eqref{weightsalphacal} from \eqref{beforekernel}.
More precisely,
given a kernel $k: \mathcal{X} \times \mathcal{X} \rightarrow \mathbb{R}$, we have $k(\mathbf{x},\mathbf{x}')= \langle\boldsymbol{\varphi}(\mathbf{x}),\boldsymbol{\varphi}(\mathbf{x}')\rangle$. 
The empirical kernel matrix $\mathbf{K} \in \mathbb{R}^{N \times N}$ is constructed as $\mathbf{K}_{i,j} = k(\mathbf{x}_i, \mathbf{x}_j)$  
and $\mathbf{k}_x \in \mathbb{R}^{N}$ is the vector defined by $\mathbf{k}_x = \begin{bmatrix}k(\mathbf{x},\mathbf{x}_1), \ldots,  k(\mathbf{x},\mathbf{x}_N)\end{bmatrix}^\top$.  

By exploiting the property of SELF, the decoder is designed such that the predictor has the form
\begin{equation}\label{decoderule}
\mathbf{s}(\mathbf{x}) = \underset{\mathbf{y}\in \mathcal{Y}}{\mathtt{argmin}}\left\langle \mathbf{c}(\mathbf{y}), V\mathbf{g}(\mathbf{x})\right\rangle_{\mathcal{H}_{\mathcal{Y}}}. 
\end{equation}

Finally, by plugging \eqref{surrofunc} into \eqref{decoderule}, we have
\begin{align}\label{consistest}
\widehat{\mathbf{s}}(\mathbf{x}) 
&= \underset{\mathbf{y}\in \mathcal{Y}}{\mathtt{argmin}}\left\langle \mathbf{c}(\mathbf{y}), V\left(\sum_{n=1}^{N}\boldsymbol{\alpha}_n(\mathbf{x})\mathbf{c}(\mathbf{y}_n)\right) \right\rangle_{\mathcal{H}_{\mathcal{Y}}} \\
&= \underset{\mathbf{y}\in \mathcal{Y}}{\mathtt{argmin}} \sum_{n=1}^{N}\boldsymbol{\alpha}_n(\mathbf{x})\Delta(\mathbf{y}, \mathbf{y}_n), \label{solu_sp}
\end{align}  
where we used the linearity property of the inner product and the definition of \eqref{defSELF} to obtain \eqref{solu_sp} from \eqref{consistest}.

In summary, when applying the implicit embedding framework to solve the structured prediction problem \eqref{probdet}, the procedure consists of two steps: 
\begin{enumerate}
	\item \textit{Surrogate learning}: Calculate the input-dependent weights $\boldsymbol{\alpha}$. 
	\item \textit{Decoding}: Optimize the $\boldsymbol{\alpha}$-weighted linear combination of losses $\Delta(\mathbf{y}, \mathbf{y}_n)$.
\end{enumerate}
The key insight is that the encoding rule $\mathbf{c}$ and the surrogate space $\mathcal{H}_{\mathcal{Y}}$ are no longer explicitly needed and in this sense the encoding is implicit.

On the basis of the implicit encoding framework~\citep{ciliberto2020general}, in the following we present the development of the proposed imitation learning algorithm.
We first outline the main idea of performing probabilistic trajectory imitation via structured prediction (Section~\ref{f-divergence}).
Afterwards, we show the strategies for trajectory modulation, which is essential for robots to reproduce motion skills in environments different from the one experienced during the demonstrations (Section~\ref{trajmod}).
Then, we illustrate motion imitation in the case of Euclidean and Riemannian probabilistic trajectories (Section~\ref{prac}).

\subsubsection{Loss Function Design by $f$-Divergence}\label{f-divergence}
We now consider addressing problem \eqref{probsetting} where the outputs are trajectory distributions in a probability space $\mathcal{P}$.  
Given an input $\mathbf{x}$, an immediate application of the solution to structured prediction \eqref{solu_sp} becomes
\begin{equation}\label{estimatorprob}
\widehat{\mathbf{s}}(\mathbf{x}) = \underset{\tilde{\mathbf{y}}\in \mathcal{P}}{\mathtt{argmin}} \sum_{n=1}^{N}\boldsymbol{\alpha}_n(\mathbf{x})\Delta(\tilde{\mathbf{y}}, \tilde{\mathbf{y}}_n),
\end{equation} 
where a suitable loss function needs to be designed to quantify the discrepancy between two probability distributions $\tilde{\mathbf{y}}$ and $\tilde{\mathbf{y}}_n$ in $ \mathcal{P}$. 
We propose to leverage tools from an information-theoretic perspective and choose the family of $f$-divergences provided the $f$-divergences generalize similarity measures between probability distributions. 
Such a choice is also in alignment with the findings from sequential decision-making that imitation learning can be treated as the divergence minimization between expert and learner trajectory distributions \citep{ke2019imitation}.

The expression for the $f$-divergence is given as\footnote{We put "true" distribution first to comply with forward KL divergence.}
\begin{equation}
	\label{f-div}
	D_f\big(\tilde{\mathbf{y}}_n(\mathbf{x}), \tilde{\mathbf{y}}(\mathbf{x})\big) \triangleq \mathbb{E}_{ \tilde{\mathbf{y}}(\mathbf{x})}\left[ f\left(\dfrac{d\tilde{\mathbf{y}}_n(\mathbf{x})}{d\tilde{\mathbf{y}}(\mathbf{x})}\right)\right],  
\end{equation} 
where $f: R^+ \rightarrow R$ represents a convex function with $f(1) = 0$.
By choosing different functions $f$, a broad class of divergences can be defined.
Common examples include the KL divergence, the reverse KL divergence, and the Jensen-Shannon divergence (see \citet{pardo2018statistical} for a full list). 
Given that different types of $f$-divergence will result in different imitation policies, we then refer to these different imitation policies as \textit{imitation modes}.

Finally, by substituting \eqref{f-div} into \eqref{estimatorprob}, the estimator for probabilistic trajectory prediction is obtained as
\begin{equation}
\label{est_div}
\widehat{\mathbf{s}}(\mathbf{x}) = \underset{\tilde{\mathbf{y}}\in \mathcal{P}}{\mathtt{argmin}} \sum_{n=1}^{N}\boldsymbol{\alpha}_n(\mathbf{x})\mathbb{E}_{ \tilde{\mathbf{y}}(\mathbf{x})}\left[ f\left(\dfrac{d\tilde{\mathbf{y}}_n(\mathbf{x})}{d\tilde{\mathbf{y}}(\mathbf{x})}\right)\right].
\end{equation}

\subsection{Trajectory Modulation}
\label{trajmod}
As the demonstration and reproduction environments can differ, robots should be able to adapt the learned motor skills during reproduction.
To this aim, it is essential to endow robots with adaptability to the arising requirements through spatial or temporal trajectory modulation. 

For example, trajectory modulation for desired \textit{via-points} can be used for reaching a region of interest or avoiding collisions.
Also, the capability of setting off from a new \textit{start point} or converging to a different \textit{end point} can make robots more flexible in tasks like pushing or pick-and-place.
Our approach guarantees that movements can be adapted on the fly during the execution of Euclidean trajectories.
Moreover, to generate more complex behaviors, multiple movement trajectories can be co-activated simultaneously \citep{duanlearning}.
Such concurrent co-activation of different trajectories is also known as trajectory \textit{superposition}, which can significantly improve motion expressiveness.

Besides spatial modulation, temporal modulation is also a necessary capability for tasks that are sensitive to correct timing. 
By speeding up or slowing down the robot's movement, trajectories can then be temporally adapted for striking-based manipulation or walking speed adjustment in locomotion.
Temporal modulation can also be used to avoid time-dependent collisions.

\subsubsection{Spatial Modulation}
We consider the issue of passing through additional desired \textit{via-points}. 
Assume that there are $J$ new desired points stored in the dataset $\mathbb{D}_v= \{\mathbf{x}_j, \tilde{\mathbf{y}}_j\}_{j=1}^J$ with each one repeating $w_j>1$ times.
To take these new requirements into account, we concatenate the new desired dataset to the original demonstrated one.
Consequently, the updated dataset to train our estimator now becomes $\mathbb{D} \cup \mathbb{D}_v$ with a total number of points 
$N' = N + \sum_j^J w_j$.
Afterwards, \eqref{estimatorprob} should be applied to the new dataset.

It is noteworthy that the size of the kernel matrix now increases to $\mathbb{R}^{N' \times N'}$.
Owing to the computational burden incurred by matrix inversion, it will be favorable to reduce the size of the kernel matrix.
To this end, we examine surrogate learning in the deterministic setting.
The optimization problem \eqref{minoptW} incorporating weighted terms is now formulated as
\begin{equation}
\underset{\mathbf{W}}{\mathtt{\min}} \frac{1}{N'}\sum_{n=1}^{N+J}w_n\|\mathbf{W}\boldsymbol{\varphi}(\mathbf{x}_n) - \mathbf{c}(\mathbf{y}_n)\|_{ \mathcal{H}_{\mathcal{Y}}}^2 + \lambda \|\mathbf{W}\|_{F}^2 
\end{equation} 
where each weight $w_n$ is defined as
\begin{equation}
w_n = \begin{cases}
w_j & \quad \mathbf{x}_n \in \mathbb{D}_v, \\
1 & \quad \mathrm{otherwise}.
\end{cases}
\end{equation}
The estimator for trajectory adaptation can still be expressed similarly to \eqref{estimatorprob}, except that the size of the kernel matrix is reduced to
$\mathbb{R}^{(N+J) \times (N+J)}$, which will yield faster computation speed when carrying out matrix inversion. 
Furthermore, the coefficients are written as
\begin{equation}\label{weightalpha}
\boldsymbol{\alpha}'(\mathbf{x}) = (\mathbf{K'}+N'\lambda \mathbf{I}_{N+J})^{-1}\mathbf{k}'_x,
\end{equation}  
where $\mathbf{K'}$ and $\mathbf{k}'_x$ are obtained by weighing the rows of $\mathbf{K}$ and $\mathbf{k}_x$ that involve $\mathbf{x}_j$ by $w_j$. 

\begin{algorithm}[t]
	\caption{Imitation Learning by Structured Prediction} 
	\label{meta_alg}
	\SetKwInput{Initialization}{Initialization}
	\SetKwInput{Trajmod}{Trajectory modulation}
	\SetKwInput{Mongen}{Motion generation}	
	\Initialization{}
	Retrieve dataset $\mathbb{D}$ from demonstrations\;
	Define kernel $k$ and hyperparameter $\lambda$\; 
	Choose imitation mode $f$\;
	\Trajmod{}
	Specify desired points in $\mathbb{D}_v$\;
    Prioritize trajectories in $\mathbb{D}_s$\;
	Aggregate dataset as $\mathbb{D} \cup \mathbb{D}_v \cup \mathbb{D}_s$ \; 
	\Mongen{}
	\textit{Input:} query point $\mathbf{x}$\;
	Calculate weights $\boldsymbol{\alpha}'(\mathbf{x})$\;
	\textit{Output:} estimated value $\widehat{\mathbf{s}}(\mathbf{x})$\;
\end{algorithm}    

As for trajectory \textit{superposition}, the robot is expected to follow $H$ prioritized trajectories. 
We denote the related dataset by $\mathbb{D}_s = \{w_h, \{\mathbf{x}_n^h, \tilde{\mathbf{y}}_n^h\}_{n=1}^N \}_{h=1}^H$ with  priorities normalized, i.e.,  $\sum_{h=1}^H w_h = 1$.
Given the assigned priorities, we construct the loss function by weighing the individual loss evaluations as:
\begin{equation}\label{weighloss}
\Delta(\tilde{\mathbf{y}}, \tilde{\mathbf{y}}_n) = \sum_{h=1}^{H}w_hD_f(\tilde{\mathbf{y}}_n^h, \tilde{\mathbf{y}}),
\end{equation}
which results in the estimator as
\begin{equation}\label{estprob}
\widehat{\mathbf{s}}(\mathbf{x}) = \underset{\tilde{\mathbf{y}}\in \mathcal{P}}{\mathtt{argmin}} \sum_{n=1}^{N}\boldsymbol{\alpha}_n(\mathbf{x})\sum_{h=1}^{H}w_hD_f(\tilde{\mathbf{y}}_n^h, \tilde{\mathbf{y}}).
\end{equation}

The algorithmic framework of the proposed imitation learning approach incorporating spatial trajectory modulation is summarized in Algorithm \ref{meta_alg}.

\subsubsection{Temporal Modulation}
When dealing with temporal modulation, a phase variable $z$ can be introduced to decouple the dependence from time \citep{ijspeert2013dynamical}.
The choice of phase $z(t)$ can be any monotonic increasing function with respect to the time stamp $t$.
By making the movement depend on the phase rather than time, a faster or slower execution of the movement is then permitted.
Therefore, the desired temporal evolution of the movement can be achieved by tuning the rate of the phase variable.
A common choice of the monotonic function is first-order linear dynamics \citep{ijspeert2013dynamical}.

\section{Practical Implementation of the Algorithm}\label{prac}
We begin by noting that Algorithm \ref{meta_alg} is a meta-algorithm since it requires solving an optimization problem over the probability space. 
In this section, we will illustrate how to practically deploy the algorithm. 
Towards this end, we make a specific choice that each output distribution satisfies a Gaussian $\tilde{\mathbf{y}}_n \sim \mathcal{N}_\alpha(\boldsymbol{\mu}_n, \boldsymbol{\Sigma}_n)$ (on a Euclidean $\mathcal{N}$ or a manifold $\mathcal{N}_\mathcal{M}$ \citep{zeestraten2017approach}) with mean $\boldsymbol{\mu}_n$ and covariance $\boldsymbol{\Sigma}_n$.
As a result, the dataset has a format of $\mathbb{D}_{\mathcal{N}} = \{\mathbf{x}_n, (\boldsymbol{\mu}_n, \boldsymbol{\Sigma}_n)\}_{n=1}^N$.  

Our intention here is to find separate estimators $\mathbf{s}_m$ and $\mathbf{s}_c$ to predict mean $\boldsymbol{\mu}$ and covariance $\boldsymbol{\Sigma}$, respectively.
Particularly, given a query point $\mathbf{x}$, the corresponding output becomes $\tilde{\mathbf{y}}(\mathbf{x}) \sim \mathcal{N}_\alpha(\mathbf{s}_m(\mathbf{x}), \mathbf{s}_c(\mathbf{x}))$.
To find these estimators, we instantiate the template of problem~\eqref{est_div} by choosing a specific $f$-divergence function and restricting the type of probabilistic trajectories to Gaussian distributions.
Consequently, problem~\eqref{est_div} becomes
\begin{equation}\label{estimatorprobGaussian}
\widehat{\mathbf{s}}(\mathbf{x}) =\!\!\! \underset{
(\boldsymbol{\mu},\boldsymbol{\Sigma} ) \in \mathcal{Y}\times  \mathcal{Y}^2
}{\mathtt{argmin}} \sum_{n=1}^{N}\boldsymbol{\alpha}_n(\mathbf{x})
D_f(\mathcal{N}_\alpha(\boldsymbol{\mu}_n, \boldsymbol{\Sigma}_n),
\mathcal{N}_\alpha(\boldsymbol{\mu}, \boldsymbol{\Sigma})).
\end{equation} 
The optimization problem \eqref{estimatorprobGaussian} can be solved by taking the derivatives with respect to the design variables $\boldsymbol{\mu}$ and $\boldsymbol{\Sigma}$, respectively, and then setting the obtained derivatives to zero. 
Therefore, we need to calculate
\begin{gather}
\sum_{n=1}^{N}\boldsymbol{\alpha}_n(\mathbf{x})\dfrac{\partial D_f(\boldsymbol{\mu}_n, \boldsymbol{\Sigma}_n, \boldsymbol{\mu}, \boldsymbol{\Sigma})}{\partial \boldsymbol{\mu}} = \mathbf{0}, \label{partialdmu} 
\\ 
\sum_{n=1}^{N}\boldsymbol{\alpha}_n(\mathbf{x})\dfrac{\partial D_f(\boldsymbol{\mu}_n, \boldsymbol{\Sigma}_n, \boldsymbol{\mu}, \boldsymbol{\Sigma})}{\partial \boldsymbol{\Sigma}} = \mathbf{0}.\label{partialdsig}
\end{gather}
When calculating the partial derivatives as required by \eqref{partialdmu} and \eqref{partialdsig}, the cost terms containing $\boldsymbol{\mu}$ and $\boldsymbol{\Sigma}$ shall be singled out, separately.
The estimators of mean $\widehat{\mathbf{s}}_m$ and covariance $\widehat{\mathbf{s}}_c$ will then appear as 
\begin{gather}\label{deltam}
\widehat{\mathbf{s}}_m(\mathbf{x}) = \underset{\boldsymbol{\mu} \in \mathcal{Y}}{\mathtt{argmin}} \sum_{n=1}^{N}\boldsymbol{\alpha}_n(\mathbf{x})\Delta_m(\boldsymbol{\mu}, \boldsymbol{\Sigma}),\\
\widehat{\mathbf{s}}_c(\mathbf{x}) = \underset{\boldsymbol{\Sigma} \in \mathcal{Y}^2}{\mathtt{argmin}} \sum_{n=1}^{N}\boldsymbol{\alpha}_n(\mathbf{x})\Delta_c(\boldsymbol{\mu}, \boldsymbol{\Sigma}), \label{deltac}
\end{gather}
where $\Delta_m$ groups all the terms containing $\boldsymbol{\mu}$ for mean prediction and $\Delta_c$ groups all the terms containing $\boldsymbol{\Sigma}$ for covariance prediction.
We drop the dependence of the cost terms on $n$ without ambiguity.
In addition, when predicting the covariance matrix, the result needs to be restricted to the cone of symmetric positive semi-definite matrices.

Next, we discuss the realization of different imitation modes by choosing different $f$-divergences in Section \ref{Euclidean}.
We show in Section \ref{manifold} how to deal with manifold-valued probabilistic trajectories.

\begin{table*}[t]
	
	\caption{List of imitation modes based on different divergences}
	\label{euclideanimitationmode}
	\centering
	\begin{tabular}{ccc}
		\toprule 
		& Kullback-Leibler divergence  &Reverse Kullback-Leibler divergence  \\ 
		
		\midrule
		$f(u)$     & $u\log(u)$                       &$-\log(u)$                    \\[1.2ex]
		$\Delta_m(\boldsymbol{\mu}, \boldsymbol{\Sigma})$ &$(\boldsymbol{\mu}- \boldsymbol{\mu}_n)^\top\boldsymbol{\Sigma}^{-1}(\boldsymbol{\mu}- \boldsymbol{\mu}_n)$ &$(\boldsymbol{\mu}- \boldsymbol{\mu}_n)^\top\boldsymbol{\Sigma}_n^{-1}(\boldsymbol{\mu}- \boldsymbol{\mu}_n)$ \\[1.2ex]
		$\widehat{\mathbf{s}}_m(\mathbf{x})$ &$\dfrac{\sum_{n=1}^{N}\boldsymbol{\alpha}_n(\mathbf{x})\boldsymbol{\mu}_n}{\sum_{n=1}^{N}\boldsymbol{\alpha}_n(\mathbf{x})}$ &$\big(\sum_{n=1}^{N}\boldsymbol{\alpha}_n(\mathbf{x})\boldsymbol{\Sigma}_n^{-1}\big)^{-1}\sum_{n=1}^{N}\boldsymbol{\alpha}_n(\mathbf{x})\boldsymbol{\Sigma}_n^{-1}\boldsymbol{\mu}_n$  \\ [2.9ex]
		$\Delta_c(\boldsymbol{\mu}, \boldsymbol{\Sigma})$ &$(\boldsymbol{\mu}- \boldsymbol{\mu}_n)^\top\boldsymbol{\Sigma}^{-1}(\boldsymbol{\mu}- \boldsymbol{\mu}_n) + \log|\boldsymbol{\Sigma}| +
		\mathtt{Tr}(\boldsymbol{\Sigma}^{-1}\boldsymbol{\Sigma}_n)$ &$-\log|\boldsymbol{\Sigma}| +
		\mathtt{Tr}(\boldsymbol{\Sigma}_n^{-1}\boldsymbol{\Sigma})$ \\[1.ex]
		$\widehat{\mathbf{s}}_c(\mathbf{x})$ &$\dfrac{\sum_{n=1}^{N}\boldsymbol{\alpha}_n(\mathbf{x})\big((\boldsymbol{\mu} - \boldsymbol{\mu}_n)(\boldsymbol{\mu}- \boldsymbol{\mu}_n)^\top+\boldsymbol{\Sigma}_n\big)}{\sum_{n=1}^{N}\boldsymbol{\alpha}_n(\mathbf{x})}$ &$\Bigg(\dfrac{\sum_{n=1}^{N}\boldsymbol{\alpha}_n(\mathbf{x})\boldsymbol{\Sigma}_n^{-1}}{\sum_{n=1}^{N}\boldsymbol{\alpha}_n(\mathbf{x})}\Bigg)^{-1}$ \\		
		\bottomrule
	\end{tabular}
\end{table*}

\subsection{Imitation with Euclidean-Valued Output}\label{Euclidean}
As discussed in Section \ref{f-divergence}, different choices of $f$ can result in different imitation strategies $\tilde{\mathbf{y}} \sim \mathcal{N}(\boldsymbol{\mu}, \boldsymbol{\Sigma})$ to imitate the expert policy $\tilde{\mathbf{y}}_n \sim \mathcal{N}(\boldsymbol{\mu}_n, \boldsymbol{\Sigma}_n)$ in response to a query input $\mathbf{x}$. 
We called these strategies imitation modes.
Here we consider the case where the trajectory mean lies in a Euclidean space.
Specifically, we show how different imitation learning algorithms can be obtained by exploiting two common $f$-divergences, namely the KL divergence and the reverse KL divergence. 

\subsubsection{KL divergence}\label{pracKL} 
We start with the well-known KL-divergence to illustrate the motivation for the cost functions design.
The KL divergence is obtained by taking  $f(u) = u\log(u)$. 
Using the properties of the KL divergence between two multivariate Gaussian distributions, the corresponding loss function is given by 
\begin{align}
&D_{\mathtt{KL}}(\tilde{\mathbf{y}}_n, \tilde{\mathbf{y}}) = \mathbb{E}_{ \tilde{\mathbf{y}}}\left[\dfrac{d\tilde{\mathbf{y}}_n}{d\tilde{\mathbf{y}}} \log\left(\dfrac{d\tilde{\mathbf{y}}_n}{d\tilde{\mathbf{y}}}\right)\right] \nonumber \\
=& \frac{1}{2}\big(\underbrace{\underbrace{(\boldsymbol{\mu}- \boldsymbol{\mu}_n)^\top\boldsymbol{\Sigma}^{-1}(\boldsymbol{\mu}- \boldsymbol{\mu}_n)}_{\textstyle\Delta_m} + \log|\boldsymbol{\Sigma}| +\mathtt{Tr}(\boldsymbol{\Sigma}^{-1}\boldsymbol{\Sigma}_n)}_{\textstyle\Delta_c} \nonumber \\
&- \log|\boldsymbol{\Sigma}_n| - \mathtt{dim}(\mathcal{Y}) \big), \label{KL-div}
\end{align}
where $ |\cdot |$ denotes the determinant of a matrix, $\mathtt{Tr}(\cdot)$ denotes the trace of a matrix, and $\mathtt{dim}(\mathcal{Y})$ indicates the dimensionality of the output space. 
 
Grouping the terms that include $\boldsymbol{\mu}$ and $\boldsymbol{\Sigma}$, respectively, we can acquire the expressions for the optimization cost functions $\Delta_m$ and $\Delta_c$ from \eqref{KL-div}.
Then we can plug these expressions in  \eqref{deltam} and \eqref{deltac} so that  the estimators with the KL-divergence imitation mode are given by
\begin{equation}\label{KLestm}
\widehat{\mathbf{s}}_m(\mathbf{x}) = \underset{\boldsymbol{\mu} \in \mathcal{Y}}{\mathtt{argmin}} \sum_{n=1}^{N}\boldsymbol{\alpha}_n(\mathbf{x})\big((\boldsymbol{\mu}- \boldsymbol{\mu}_n)^\top\boldsymbol{\Sigma}^{-1}(\boldsymbol{\mu}- \boldsymbol{\mu}_n)\big),
\end{equation}
\begin{multline}\label{KLestc}
\widehat{\mathbf{s}}_c(\mathbf{x}) = \underset{\boldsymbol{\Sigma} \in \mathcal{Y}^2}{\mathtt{argmin}} \sum_{n=1}^{N}\boldsymbol{\alpha}_n(\mathbf{x})\big((\boldsymbol{\mu}- \boldsymbol{\mu}_n)^\top\boldsymbol{\Sigma}^{-1}(\boldsymbol{\mu}- \boldsymbol{\mu}_n) \\ + \log|\boldsymbol{\Sigma}| +
\mathtt{Tr}(\boldsymbol{\Sigma}^{-1}\boldsymbol{\Sigma}_n)\big).
\end{multline} 
To compute the optimal mean and covariance predictions, we set to zero the derivatives of \eqref{KLestm} and \eqref{KLestc} with respect to $\boldsymbol{\mu}$ and $\boldsymbol{\Sigma}$, which yields
\begin{equation}\label{KLoptm}
\boldsymbol{\mu} = \dfrac{\sum_{n=1}^{N}\boldsymbol{\alpha}_n(\mathbf{x})\boldsymbol{\mu}_n}{\sum_{n=1}^{N}\boldsymbol{\alpha}_n(\mathbf{x})},
\end{equation}
\begin{align}
\boldsymbol{\Sigma} &= \frac{\sum_{n=1}^{N}\boldsymbol{\alpha}_n(\mathbf{x})\big((\boldsymbol{\mu}- \boldsymbol{\mu}_n)(\boldsymbol{\mu}- \boldsymbol{\mu}_n)^\top+\boldsymbol{\Sigma}_n\big)}{\sum_{n=1}^{N}\boldsymbol{\alpha}_n(\mathbf{x})}\label{KLoptco}\\
&\approx \frac{\sum_{n=1}^{N}\boldsymbol{\alpha}_n(\mathbf{x})\boldsymbol{\Sigma}_n}{\sum_{n=1}^{N}\boldsymbol{\alpha}_n(\mathbf{x})}.\label{KLoptcoapp}
\end{align}

We note that the prediction of the covariance matrix according to \eqref{KLoptco} is dependent on the predicted mean value given by \eqref{KLoptm}. 
By contrast, the prediction of the mean value is only dependent on the training output mean $\boldsymbol{\mu}_n$.
Therefore, we should predict the mean first before computing the covariance prediction.
In addition, if we would like to alleviate the interference from the mean value on the covariance prediction, it can be considered to approximate~\eqref{KLoptco} by omitting the term $(\boldsymbol{\mu}- \boldsymbol{\mu}_n)(\boldsymbol{\mu}- \boldsymbol{\mu}_n)^\top$ as in~\eqref{KLoptcoapp}.

\subsubsection{Reverse KL divergence}
Besides the KL divergence, there are plenty of other divergences that can be used to construct loss functions. 
For example, we here consider the reverse KL divergence, which reflects the asymmetry of the KL divergence.
The reverse KL divergence is defined by $f(u) = -\log(u)$, which gives rise to the associated loss function as
\begin{align}\label{RKL-div}
&D_{\mathtt{RKL}}(\tilde{\mathbf{y}}_n, \tilde{\mathbf{y}}) = \mathbb{E}_{ \tilde{\mathbf{y}}}\left[ \log\left(\dfrac{d\tilde{\mathbf{y}}}{d\tilde{\mathbf{y}}_n}\right)\right]\nonumber\\
=&\frac{1}{2}\big(\underbrace{(\boldsymbol{\mu}-\boldsymbol{\mu}_n)^\top\boldsymbol{\Sigma}_n^{-1}(\boldsymbol{\mu}-\boldsymbol{\mu}_n)}_{\textstyle\Delta_m} \underbrace{-\log|\boldsymbol{\Sigma}| + \mathtt{Tr}(\boldsymbol{\Sigma}_n^{-1}\boldsymbol{\Sigma})}_{\textstyle\Delta_c} \nonumber\\
&+ \log|\boldsymbol{\Sigma}_n|-
\mathtt{dim}(\mathcal{Y}) \big).
\end{align}

Similarly to Section \ref{pracKL}, we collect all the terms involving $\boldsymbol{\mu}$ to specify the cost function $\Delta_m$ for mean prediction and collect all the terms involving $\boldsymbol{\Sigma}$ to specify the cost function $\Delta_c$ for covariance prediction.
By plugging the cost functions given by \eqref{RKL-div} into \eqref{deltam} and \eqref{deltac}, the reverse-KL imitation mode estimators can be expressed as
\begin{gather}\label{rkl_estm}
\widehat{\mathbf{s}}_m(\mathbf{x}) = \underset{\boldsymbol{\mu} \in \mathcal{Y}}{\mathtt{argmin}} \sum_{n=1}^{N}\boldsymbol{\alpha}_n(\mathbf{x})\big((\boldsymbol{\mu}- \boldsymbol{\mu}_n)^\top\boldsymbol{\Sigma}_n^{-1}(\boldsymbol{\mu}- \boldsymbol{\mu}_n)\big),\\
\widehat{\mathbf{s}}_c(\mathbf{x}) = \underset{\boldsymbol{\Sigma} \in \mathcal{Y}^2}{\mathtt{argmin}} \sum_{n=1}^{N}\boldsymbol{\alpha}_n(\mathbf{x})\big(-\log|\boldsymbol{\Sigma}| +
\mathtt{Tr}(\boldsymbol{\Sigma}_n^{-1}\boldsymbol{\Sigma})\big).\label{rkl_estc}
\end{gather}

The solutions are also calculated by setting the derivatives of \eqref{rkl_estm} and \eqref{rkl_estc} with respect to the design variables $\boldsymbol{\mu}$ and $\boldsymbol{\Sigma}$ equal to zero, respectively.
Therefore, the optimal predictions for mean and covariance are given by 
\begin{gather}\label{RKLm}
\boldsymbol{\mu} = 
\left(\sum_{n=1}^{N}\boldsymbol{\alpha}_n(\mathbf{x})\boldsymbol{\Sigma}_n^{-1}\right)^{-1}\sum_{n=1}^{N}\boldsymbol{\alpha}_n(\mathbf{x})\boldsymbol{\Sigma}_n^{-1}\boldsymbol{\mu}_n,\\
\boldsymbol{\Sigma} = \left(\frac{\sum_{n=1}^{N}\boldsymbol{\alpha}_n(\mathbf{x})\boldsymbol{\Sigma}_n^{-1}}{\sum_{n=1}^{N}\boldsymbol{\alpha}_n(\mathbf{x})}\right)^{-1}. \label{RKLcov}
\end{gather}

It can be seen that the prediction of the mean value by~\eqref{RKLm} relies on both the training output covariance $\boldsymbol{\Sigma}_n$ and mean $\boldsymbol{\mu}_n$.
By contrast, the prediction of the covariance matrix in~\eqref{RKLcov} only depends on the training output covariance.
As a side note, the predicted covariance matrix given by~\eqref{RKLcov} can be utilized for mean prediction to avoid repetitive computation of the term $(\sum_{n=1}^{N}\boldsymbol{\alpha}_n(\mathbf{x})\boldsymbol{\Sigma}_n^{-1})^{-1}$. 

\begin{algorithm}[t]
	\caption{Imitation with Euclidean-Valued Output} 
	\label{imit_Euclidean}
	Collect trajectories from multiple demonstrations\; 
	\tcc{Assumption of Gaussian output}
	Process raw data for $\mathbb{D}_{\mathcal{N}} = \{\mathbf{x}_n, (\boldsymbol{\mu}_n, \boldsymbol{\Sigma}_n)\}_{n=1}^N$\;
	Define the kernel $k$ and parameter $\lambda$\;	   
	Choose imitation mode $f$\;
	\Switch{imitation mode $f$}{
		\For{$\mathbf{x} = \mathbf{x}_{\mathrm{Start}}, \ldots, \mathbf{x}_{\mathrm{End}}$}{
			\textit{Input:} a query point $\mathbf{x}$\;
			Calculate the weights $\boldsymbol{\alpha}(\mathbf{x})$\;
			\Case{$f(u) = u\log(u)$}{
				\tcc{Predict with the KL divergence mode}
				\textit{Output:} mean $\boldsymbol{\mu}$ as per \eqref{KLoptm}\;
				\textit{Output:} covariance $\boldsymbol{\Sigma}$ as per \eqref{KLoptco}\;
			}
			\Case{$f(u) = -\log(u)$}{
				\tcc{Predict with the reverse KL divergence mode}
				\textit{Output:} mean $\boldsymbol{\mu}$ as per \eqref{RKLm}\;
				\textit{Output:} covariance $\boldsymbol{\Sigma}$ as per \eqref{RKLcov}\;
			}
			\Other{
				Formulate $\Delta_m$ and $\Delta_c$ motivated by \eqref{estimatorprobGaussian}\;
				\textit{Output:} mean $\boldsymbol{\mu}$ by optimizing \eqref{deltam}\;
				\textit{Output:} covariance $\boldsymbol{\Sigma}$ by optimizing \eqref{deltac}\;\tcp{s.t. $\boldsymbol{\Sigma} = \boldsymbol{\Sigma}^\top$ and $\boldsymbol{\Sigma} \succeq 0$} 
			}
		}
	}
\end{algorithm}

As a sanity check, it can be seen that the predicted covariance matrix in \eqref{KLoptco} and \eqref{RKLcov} is indeed symmetric positive semi-definite if the training covariance matrices are symmetric positive definite
since (\romannumeral 1) the outer product of a vector, (\romannumeral 2) the inverse of a symmetric positive definite matrix, and (\romannumeral 3) the sum of the symmetric positive semi-definite matrices are all symmetric positive semi-definite matrices. 
While the weight $\boldsymbol{\alpha}_n(\mathbf{x})$ can be non-positive for some terms, in practice, its score at the proximity to the input query point is positive and will usually dominate other terms.
For other possible choices of $f$, the constraint for $\boldsymbol{\Sigma}$ being a symmetric positive semi-definite matrix should be taken into account explicitly.

It should be noted that the optimization problem \eqref{estimatorprobGaussian} is derived on the basis of the implicit embedding framework.
Therefore, to make the formulation valid, it is important for the divergence-inspired loss functions $D_{\mathtt{KL}}$ \eqref{KL-div} and $D_{\mathtt{RKL}}$ \eqref{RKL-div} to be SELF. 
In fact, it is possible to show these are the cases.
We provide a proof sketch in Appendix \ref{appSELF}.
 
Our main results are summarized in Table \ref{euclideanimitationmode} and the algorithm for imitation of Euclidean-valued probabilistic trajectories is written in Algorithm \ref{imit_Euclidean}.

\subsubsection{Learning the Velocity Profile of Temporal Trajectories}\label{veladapt}
In tasks such as throwing and catching, robots need to exhibit dynamic behavior, which requires learning motor skills involving not only position but also velocity or even higher-order derivatives.
However, a direct application of the structured prediction method developed so far could be infeasible, since both mean prediction strategies given by \eqref{KLoptm} and \eqref{RKLm} are not aware of the constraint on the derivative relationship between position and velocity outputs.  
Therefore, we need to explicitly cope with the issue of learning temporal trajectories.
We denote the output value $\mathbf{y}$ composed of both position $\boldsymbol{\mu}$ and velocity $\boldsymbol{\dot{\mu}}$ as
\begin{equation}\label{outputtemporalstructure}
\mathbf{y}(t) = 
\begin{bmatrix} \boldsymbol{\mu}(t) \\
\boldsymbol{\dot{\mu}}(t) 
\end{bmatrix}.
\end{equation}

Recall that we used the least-squares estimator \eqref{minoptW} to solve the surrogate problem.
As this treatment can not take the underlying temporal constraint of output values into account, a derivative constraint-aware estimator $\mathbf{g}$ is needed such that it can capture the temporal relationship.    

We motivate our design for the estimation value $\widehat{\mathbf{g}}$ by referring to an analytic insight of the feature map.
Specifically, to complement \eqref{gpara} for velocity learning, we employ the following parametric form
\begin{equation}\label{temporalparametricform}
\mathbf{g}(t) = \begin{bmatrix} \mathbf{W}^\top\boldsymbol{\varphi}(t) \\
\mathbf{W}^\top\boldsymbol{\dot{\varphi}}(t)
\end{bmatrix}
\end{equation}
where the same weight matrix $\mathbf{W}$ is used for both velocity and position feature maps so that the time derivative constraint can be respected.
The feature maps for velocity learning are expressed by
\begin{equation}\label{defphidot}
\boldsymbol{\dot{\varphi}}(t) = \lim_{\delta\to 0} \dfrac{\boldsymbol{\varphi}(t+\delta)-\boldsymbol{\varphi}(t-\delta)}{2\delta}.
\end{equation}
The estimated value for $\mathbf{W}$ can be obtained by solving the following optimization problem 
\begin{equation}\label{optW}
\underset{\mathbf{W}}{\min} \; \dfrac{1}{N}\sum_{n=1}^N \left\|\begin{bmatrix}
\boldsymbol{\varphi}_n^\top\\ \boldsymbol{\dot{\varphi}}_n^\top 
\end{bmatrix}\mathbf{W} - \begin{bmatrix}
\boldsymbol{\mu}_n^\top\\ \boldsymbol{\dot{\mu}}_n^\top 
\end{bmatrix} \right\|^2_F + \lambda \|\mathbf{W}\|^2_F.
\end{equation}
By setting the derivative of \eqref{optW} with respect to $\mathbf{W}$ to zero, we then obtain
\begin{equation}\label{whattemporal}
\widehat{\mathbf{W}} = \boldsymbol{\Phi}(\boldsymbol{\Phi}^\top\boldsymbol{\Phi}+N\lambda\mathbf{I})^{-1}\mathbf{Y},
\end{equation}
where we represent $\boldsymbol{\Phi} = \begin{bmatrix}\boldsymbol{\varphi}_1, \boldsymbol{\dot{\varphi}}_1,\ldots,\boldsymbol{\varphi}_N, \boldsymbol{\dot{\varphi}}_N\end{bmatrix}$
and 
$\mathbf{Y} = \begin{bmatrix}\boldsymbol{\mu}_1,\boldsymbol{\dot{\mu}}_1, \ldots, \boldsymbol{\mu}_N,\boldsymbol{\dot{\mu}}_N\end{bmatrix}^\top$.
Consequently, by substituting \eqref{whattemporal} into \eqref{temporalparametricform}, we have the predicted values at time step $t$ as
\begin{equation}\label{yhatsol}
\widehat{\mathbf{g}}(t) = \begin{bmatrix} \mathbf{Y}^\top(\mathbf{K}+N\lambda\mathbf{I})^{-1}\mathbf{k}^p \\
\mathbf{Y}^\top(\mathbf{K}+N\lambda\mathbf{I})^{-1}\mathbf{k}^v
\end{bmatrix}
\end{equation}
where the vectors $\mathbf{k}^p$ and $\mathbf{k}^v$ are given by
\begin{equation}\label{defkvkp}
\mathbf{k}^p_i = \begin{cases} \boldsymbol{\varphi}_i^\top\boldsymbol{\varphi}(t)\\\boldsymbol{\dot{\varphi}}_i^\top\boldsymbol{\varphi}(t) \end{cases} 
\mathbf{k}^v_i = \begin{cases}  \boldsymbol{\varphi}_i^\top\boldsymbol{\dot{\varphi}}(t) & i=2n-1,\\\boldsymbol{\dot{\varphi}}_i^\top\boldsymbol{\dot{\varphi}}(t) & i=2n.\end{cases}
\end{equation}
The kernel matrix $\mathbf{K}$ is constructed as
\begin{equation}\label{defKphi}
\mathbf{K}_{i,j} = \begin{cases}\boldsymbol{\varphi}_i^\top\boldsymbol{\varphi}_j & \quad i=2n-1, \quad j=2n'-1,\\
\boldsymbol{\varphi}_i^\top\boldsymbol{\dot{\varphi}}_j & \quad i=2n-1,\quad j=2n',\\
\boldsymbol{\dot{\varphi}}_i^\top\boldsymbol{\varphi}_j & \quad i=2n, \quad j=2n'-1, \\
\boldsymbol{\dot{\varphi}}_i^\top\boldsymbol{\dot{\varphi}}_j & \quad i=2n, \quad j=2n',
\end{cases}
\end{equation}
where $n$ and $n' = 1,2,3, \ldots, N$. 
After substituting $\boldsymbol{\dot{\varphi}}$ with its definition \eqref{defphidot} into \eqref{defKphi}, we apply the kernel trick for the matrix as also shown by \citep{huang2019kernelized}.
In summary, the definition for the kernel matrix is given in Table \ref{defK}, where $t^{\pm} \triangleq t\pm \delta$. 
Also, the kernelized version for \eqref{defkvkp} can be derived similarly.

\begin{table}[h]
	\setlength{\tabcolsep}{6pt} 
	\renewcommand{\arraystretch}{1.8} 
	\small\sf\centering
	\caption{Definition for kernel matrix $\mathbf{K}$. \label{defK}}
	\begin{tabular}{c||cc}
		\hline
		$\mathbf{K}_{i,j}$&$i=2n-1$&$i=2n$\\
		\hline 
		$j=2n'-1$&$k(t_i,t_j)$&$(k(t_i^+,t_j)-k(t_i^-,t_j))/2\delta$\\ \hline
		$j=2n'$&\begin{tabular}{@{}c@{}}$(k(t_i,t_j^+)-$\\[-5pt]$k(t_i,t_j^-))/2\delta$\end{tabular}&\begin{tabular}{@{}c@{}} $(k(t_i^+,t_j^+)-k(t_i^+,t_j^-)-$ \\[-5pt]
			$k(t_i^-,t_j^+)+k(t_i^-,t_j^-))/4\delta^2$ 
		\end{tabular}\\
		\hline
	\end{tabular}
\end{table}

It should be noted that despite the presence of the time derivative constraint in \eqref{outputtemporalstructure}, it still admits the Euclidean metric for such output values.
Hence, there is no need for an encoding rule, as evident by \eqref{optW} where the temporal trajectory outputs are directly used as the learning objectives instead of their embedding values. 
With the collapse of the surrogate space, the structured prediction estimator $\widehat{\mathbf{s}}(t)$ thus coincides with the mapping of surrogate learning \eqref{yhatsol}.  

Furthermore, to align with the formation of \eqref{KLoptm}, we re-write the solution to our temporal estimator $\widehat{\mathbf{s}}(t)$ towards the formalism of a weighted sum of output values:
\begin{equation}
\widehat{\mathbf{s}}(t) = \sum_{n=1}^N \boldsymbol{\alpha}_n\mathbf{y}_n.
\end{equation}
The weight matrix $\boldsymbol{\alpha}_n$ is defined as  
\begin{equation}
\boldsymbol{\alpha}_n = \begin{bmatrix}
\boldsymbol{\alpha}^{p}_{2n-1}\mathbf{I} & \boldsymbol{\alpha}^{p}_{2n}\mathbf{I}\\
\boldsymbol{\alpha}^{v}_{2n-1}\mathbf{I} & \boldsymbol{\alpha}^{v}_{2n}\mathbf{I}
\end{bmatrix}
\end{equation}
where we have
\begin{equation}
\boldsymbol{\alpha}^{p} = (\mathbf{K}+\lambda\mathbf{I})^{-1}\mathbf{k}^p 
\;\, \mathrm{and} \,\; 
\boldsymbol{\alpha}^{v} = (\mathbf{K}+\lambda\mathbf{I})^{-1}\mathbf{k}^v.    
\end{equation}

To modulate the temporal trajectory such that it can pass through some desired via-point and/or via-velocity, $\mathbf{K}$, $\mathbf{k}^p$, and $\mathbf{k}^v$ need to be modified similarly to \eqref{weightalpha}, i.e., the rows that involve desired adaptive behavior need to be weighed accordingly.

\subsection{Imitation with Manifold-Valued Output}\label{manifold}
In this section, we present our algorithmic framework in the context of manifold structured prediction, i.e., when the output space $\mathcal{Y}$ is a manifold $\mathcal{M}$.     
Differently from Section \ref{Euclidean}, the learner's and the expert's probabilistic trajectories are now constrained to lie on a Riemannian manifold\footnote{Basic notions and nomenclatures on Riemannian manifolds are recalled in Appendix \ref{Rie}.}.
In this case, we consider the Riemannian Gaussian $\mathcal{N}_{\mathcal{M}}$ to represent the policy, which is usually approximated as (see e.g., \citet{simo20173d,zeestraten2017approach}):
\begin{equation}
\mathcal{N}_{\mathcal{M}}(\mathbf{y}; \boldsymbol{\mu}, \boldsymbol{\Sigma}) = \frac{1}{\sqrt{(2\pi)^d|\boldsymbol{\Sigma}|}}e^{-\frac{1}{2}\mathtt{Log}_{\boldsymbol{\mu}}(\mathbf{y})^\top\boldsymbol{\Sigma}^{-1}\mathtt{Log}_{\boldsymbol{\mu}}(\mathbf{y})}
\end{equation}
where $\boldsymbol{\mu} \in \mathcal{M}$ is the Riemannian mean and $\boldsymbol{\Sigma}$ is the covariance matrix defined in the tangent space $\mathcal{T}_{\boldsymbol{\mu}}\mathcal{M}$. 

To construct the corresponding loss function, we also propose to compute the $f$-divergence between the learner policy $\tilde{\mathbf{y}} \sim \mathcal{N}_{\mathcal{M}}(\boldsymbol{\mu},\boldsymbol{\Sigma})$ and the expert policy $\tilde{\mathbf{y}}_{n} \sim \mathcal{N}_{\mathcal{M}}(\boldsymbol{\mu}_{n},\boldsymbol{\Sigma}_{n})$.
Here, we take the KL divergence as an example, so that 
the loss function is  
\begin{align}\label{manifoldKL}
&D_{\mathtt{KL}}(\tilde{\mathbf{y}}_n, \tilde{\mathbf{y}}) 
= \frac{1}{2} \int \bigg(\log\dfrac{|\boldsymbol{\Sigma}|}{|\boldsymbol{\Sigma}_n|} -\mathtt{Log}_{\boldsymbol{\mu}_n}(\mathbf{y})\boldsymbol{\Sigma}_n^{-1}\mathtt{Log}_{\boldsymbol{\mu}_n}(\mathbf{y}) \nonumber\\ 
&\; \quad \quad \quad \quad 
+\mathtt{Log}_{\boldsymbol{\mu}}(\mathbf{y})\boldsymbol{\Sigma}^{-1}\mathtt{Log}_{\boldsymbol{\mu}}(\mathbf{y})\bigg )\tilde{\mathbf{y}}_n \,d\mathbf{y}.
\end{align}

To ease  the computation of \eqref{manifoldKL}, we make use of the following approximation
\begin{equation}
\mathtt{Log}_{\boldsymbol{\mu}}(\mathbf{y}) \approx \mathtt{Log}_{\boldsymbol{\mu}}(\boldsymbol{\mu}_n) + \mathtt{Log}_{\boldsymbol{\mu}_n}(\mathbf{y}).
\end{equation}
As a result, \eqref{manifoldKL} can be approximated as
\begin{align}
&\frac{1}{2}\big(\underbrace{\underbrace{\mathtt{Log}_{\boldsymbol{\mu}}(\boldsymbol{\mu}_n)^\top\boldsymbol{\Sigma}^{-1}\mathtt{Log}_{\boldsymbol{\mu}}(\boldsymbol{\mu}_n)}_{\textstyle\Delta_m} + \log|\boldsymbol{\Sigma}| +\mathtt{Tr}(\boldsymbol{\Sigma}^{-1}\boldsymbol{\Sigma}_n)}_{\textstyle\Delta_c} \nonumber \\
&- \log|\boldsymbol{\Sigma}_n| - \mathtt{dim}(\mathcal{Y}) \big). \label{KL-manifold}
\end{align}

In view of the complex form of $\textstyle\Delta_m$, we propose to ease the computations by omitting the weight $\boldsymbol{\Sigma}$ upon observing that it plays no role in the Euclidean case \eqref{KLoptm}.
Therefore, the cost function for mean prediction $\textstyle\Delta_m$ now becomes
\begin{equation}
\Delta_m = \mathtt{Log}_{\boldsymbol{\mu}}(\boldsymbol{\mu}_n)^\top\mathtt{Log}_{\boldsymbol{\mu}}(\boldsymbol{\mu}_n) \approx \mathtt{dist}^2(\boldsymbol{\mu}_n,\boldsymbol{\mu}),
\end{equation}
where $\mathtt{dist}(\cdot, \cdot)$ denotes the geodesic distance between two manifold points.
Finally, our estimator for mean prediction is given by 
\begin{equation}\label{manifoldestm}
\widehat{\mathbf{s}}_m(\mathbf{x}) = \underset{\boldsymbol{\mu} \in \mathcal{M}}{\mathtt{argmin}} \sum_{n=1}^{N}\boldsymbol{\alpha}_n(\mathbf{x})\; \mathtt{dist}^2(\boldsymbol{\mu}_n, \boldsymbol{\mu}).
\end{equation}

\begin{table*}[t]
	\small\sf\centering
	\caption{List of operations of common Riemannian manifolds considered in this paper.\label{T1}}
	\begingroup
	\setlength{\tabcolsep}{1.5pt} 
	\renewcommand{\arraystretch}{1} 
	\begin{tabular}{lcc}
		\toprule
		&Sphere with radius $r$: $\mathbb{S}^2(r)$ &Circular generalized cylinder: $\mathbb{R}^2\times\mathbb{S}^1$ \\
		\midrule 
		\begin{tabular}{@{}c@{}}Distance metric\\ $\mathtt{dist}(\boldsymbol{\mu}_n, \boldsymbol{\mu})$ \end{tabular} &$r \arccos\left(\dfrac{\boldsymbol{\mu}_n^\top\boldsymbol{\mu}}{r^2}\right)$ & $\sqrt{\left\|\boldsymbol{\mu}_n^{\mathbb{R}}-\boldsymbol{\mu}^{\mathbb{R}}\right\|^2+\arccos\left({\boldsymbol{\mu}_n^{\mathcal{S}}}^\top\boldsymbol{\mu}^{\mathcal{S}}\right)^2}$ \\ [3ex]
		\begin{tabular}{@{}c@{}}Minimization objective\\ $\mathtt{F}(\boldsymbol{\mu})$ \end{tabular} &$\sum_{n=1}^{N}\boldsymbol{\alpha}_nr^2\arccos\left(\dfrac{\boldsymbol{\mu}_n^\top\boldsymbol{\mu}}{r^2}\right)^2$ &$\sum_{n=1}^{N}\boldsymbol{\alpha}_n\left(\left\|\boldsymbol{\mu}_n^{\mathbb{R}}-\boldsymbol{\mu}^{\mathbb{R}}\right\|^2+\arccos\left({\boldsymbol{\mu}_n^{\mathcal{S}}}^\top\boldsymbol{\mu}^{\mathcal{S}}\right)^2\right)$ \\[3ex]
		\begin{tabular}{@{}c@{}}Riemannian gradient \\
			$\nabla_{\mathcal{M}} \mathtt{F}(\boldsymbol{\mu})$ \end{tabular} &$2\sum_{n=1}^{N}\boldsymbol{\alpha}_n\left(\dfrac{\boldsymbol{\mu}\boldsymbol{\mu}^{\top}}{r^2}-\mathbf{I}\right)\dfrac{\arccos\left(\dfrac{\boldsymbol{\mu}_n^\top\boldsymbol{\mu}}{r^2}\right)}{\sqrt{1-\left(\dfrac{\boldsymbol{\mu}_n^\top\boldsymbol{\mu}}{r^2}\right)^2}}\boldsymbol{\mu}_n$ &\begin{tabular}{@{}r@{}} $2\sum_{n=1}^{N}\boldsymbol{\alpha}_n\Bigg[\left[\boldsymbol{\mu}^{\mathbb{R}}-\boldsymbol{\mu}_n^{\mathbb{R}}\right]^\top \, \dfrac{\arccos\left({\boldsymbol{\mu}_n^{\mathcal{S}}}^\top\boldsymbol{\mu}^{\mathcal{S}}\right)}{\sqrt{1-\left({\boldsymbol{\mu}_n^{\mathcal{S}}}^\top\boldsymbol{\mu}^{\mathcal{S}}\right)^2}}$\\$\times\left[(\boldsymbol{\mu}^{\mathcal{S}}{\boldsymbol{\mu}^{\mathcal{S}}}^{\top}-\mathbf{I})\boldsymbol{\mu}_n^{\mathcal{S}}\right]^\top\Bigg]^\top$\end{tabular} \\[6.ex]
		Retraction $R_{\boldsymbol{\mu}}(\mathfrak{p})$ &$r \dfrac{\boldsymbol{\mu} + \mathfrak{p}}{\|\boldsymbol{\mu} + \mathfrak{p}\|}$  &$\left[\left[\boldsymbol{\mu}^{\mathbb{R}}+\mathfrak{p}^{\mathbb{R}}\right]^\top \, \left[\dfrac{\boldsymbol{\mu}^{\mathcal{S}} + \mathfrak{p}^{\mathcal{S}}}{\|\boldsymbol{\mu}^{\mathcal{S}} + \mathfrak{p}^{\mathcal{S}}\|}\right]^\top\right]^\top$ \\[3ex]
		Logarithmic map $\mathtt{Log}_{\boldsymbol{\mu}_n}(\boldsymbol{\mu})$ &$\mathtt{dist}(\boldsymbol{\mu}_n, \boldsymbol{\mu})\dfrac{r^2\boldsymbol{\mu}-\boldsymbol{\mu}_n^\top\boldsymbol{\mu}\boldsymbol{\mu}_n}{\|r^2\boldsymbol{\mu}-\boldsymbol{\mu}_n^\top\boldsymbol{\mu}\boldsymbol{\mu}_n\|}$ 
		&$\left[\left[\boldsymbol{\mu}^{\mathbb{R}}-\boldsymbol{\mu}^{\mathbb{R}}_n \right]^\top \, \mathtt{dist}(\boldsymbol{\mu}_n^{\mathcal{S}}, \boldsymbol{\mu}^{\mathcal{S}})\dfrac{{\boldsymbol{\mu}^{\mathcal{S}}}^\top\!\!-\boldsymbol{\mu}_n^\top\boldsymbol{\mu}{\boldsymbol{\mu}_n^{\mathcal{S}}}^\top}{\|\boldsymbol{\mu}^{\mathcal{S}}-\boldsymbol{\mu}_n^\top\boldsymbol{\mu}\boldsymbol{\mu}_n^{\mathcal{S}}\|}\right]^\top$ \\[3ex]
		\begin{tabular}{@{}c@{}}Parallel transport\\ $\Gamma_{\boldsymbol{\mu}_n \to \boldsymbol{\mu}}(\mathfrak{u})$\end{tabular} 
		&$\mathfrak{u} \!-\!\! \dfrac{\mathtt{Log}_{\boldsymbol{\mu}_n}(\boldsymbol{\mu})^\top\mathfrak{u}}{\mathtt{dist}^2(\boldsymbol{\mu}_n, \boldsymbol{\mu})}\bigl(\mathtt{Log}_{\boldsymbol{\mu}_n}\boldsymbol{\mu} \!+\! \mathtt{Log}_{\boldsymbol{\mu}}\boldsymbol{\mu}_n \bigr)$ 
		& $\left[{\mathfrak{u}^{\mathbb{R}}}^{\!\top} \, {\mathfrak{u}^{\mathcal{S}}}^\top \!\!\!-\!\! \dfrac{\mathtt{Log}_{\boldsymbol{\mu}_n^{\mathcal{S}}}(\boldsymbol{\mu}^{\mathcal{S}})^{\!\top}\!\mathfrak{u}^{\mathcal{S}}}{\mathtt{dist}^2(\boldsymbol{\mu}_n^{\mathcal{S}}, \boldsymbol{\mu}^{\mathcal{S}})}\!\!
		\left[\mathtt{Log}_{\boldsymbol{\mu}_n^{\mathcal{S}}}\boldsymbol{\mu}^{\mathcal{S}} \!\!+\! \mathtt{Log}_{\boldsymbol{\mu}^{\mathcal{S}}}\boldsymbol{\mu}_n^{\mathcal{S}} \right]^{\!\top} \right]^{\!\top}$\\
		\bottomrule
	\end{tabular}
	\endgroup
\end{table*}

\begin{algorithm}[t]
	\caption{Imitation with Manifold Output} 
	\label{manifold_alg}
	Collect multiple Riemannian trajectories\;
	Process raw data for $\mathbb{D}_{\mathcal{M}} = \{\mathbf{x}_n, (\boldsymbol{\mu}_n, \boldsymbol{\Sigma}_n)\}_{n=1}^N$\;
	Eigen-decomposition of covariance as per \eqref{eigendecomp}\; 
	Initialize kernel $k$, regularization $\lambda$, and step size $\eta$\;
	\For{$\mathbf{x} = \mathbf{x}_{\mathrm{Start}}, \ldots, \mathbf{x}_{\mathrm{End}}$}{
		\textit{Input:} a query point $\mathbf{x}$\;
		Calculate the weights $\boldsymbol{\alpha}(\mathbf{x})$\;
		\Repeat{convergence}{
			$\mathfrak{v} =\nabla_{\mathcal{M}}\sum_{n=1}^{N}\boldsymbol{\alpha}_n(\mathbf{x})\; \mathtt{dist}^2(\boldsymbol{\mu}_n, \boldsymbol{\mu})$\;
			$\boldsymbol{\mu} \leftarrow R_{\boldsymbol{\mu}}(\eta\,\mathfrak{v})$\;
		}	
		\textit{Output:} mean $\boldsymbol{\mu}$\;
		Compute the parallel transport covariance $\boldsymbol{\Sigma}_{\parallel n}$ as per \eqref{parallel_cov}\;
		\textit{Output:} covariance $\boldsymbol{\Sigma}$ as per \eqref{manifoldco}\;
	}
\end{algorithm}
We note that the loss function used in \eqref{manifoldestm} is the squared geodesic distance, which is SELF as shown by \citet{rudi2018manifold}.
To perform the estimation at a new test point $\mathbf{x}$, geometric optimization is required. 
In particular, we consider the Riemannian gradient descent, which extends the usual gradient descent method to manifolds with the guarantee that the computed value is still an element of the manifold. 

By denoting the minimization objective of \eqref{manifoldestm} as $\mathtt{F}(\boldsymbol{\mu})$, the iterative optimization process takes the form
\begin{equation}
\boldsymbol{\mu}_{i+1} = \mathtt{Exp}_{\boldsymbol{\mu}_i}\left(\eta_i \nabla_{\mathcal{M}} \mathtt{F}(\boldsymbol{\mu}_i)\right),
\end{equation}
where $\nabla_{\mathcal{M}}$ is the Riemannian gradient operator and $\eta_i \in \mathbb{R}$ is a step size.
Since the exponential map $\mathtt{Exp}$ could be difficult to compute, a computationally cheaper alternative is the retraction map $R_{\boldsymbol{\mu}}: \mathcal{T}_{\boldsymbol{\mu}}\mathcal{M} \rightarrow \mathcal{M}$, which is a first-order approximation to the exponential map.
In addition to faster computation, the retraction map also guarantees convergence.

Apart from the Riemannian gradient descent, in some simple cases, it is also possible to perform Riemannian optimization by optimizing in the Euclidean space first and then projecting the results onto the manifold.
However, for complex manifolds, the projection operation can be very expensive to compute.
By contrast, our employed Riemannian gradient descent provides a principled way for Riemannian optimization by making use of the intrinsic geometry of manifolds.

The procedure for the covariance matrix prediction can be achieved similarly to \eqref{KLoptco}:
\begin{align}\label{manifoldco}
\boldsymbol{\Sigma} &= \frac{\sum_{n=1}^{N}\boldsymbol{\alpha}_n(\mathbf{x})\big(\mathtt{Log}_{\boldsymbol{\mu}}(\boldsymbol{\mu}_n)\mathtt{Log}_{\boldsymbol{\mu}}(\boldsymbol{\mu}_n)^\top+\boldsymbol{\Sigma}_{\parallel n}\big)}{\sum_{n=1}^{N}\boldsymbol{\alpha}_n(\mathbf{x})},\\
&\approx \frac{\sum_{n=1}^{N}\boldsymbol{\alpha}_n(\mathbf{x})\boldsymbol{\Sigma}_{\parallel n}}{\sum_{n=1}^{N}\boldsymbol{\alpha}_n(\mathbf{x})},
\end{align}
where a parallel transported covariance matrix $\boldsymbol{\Sigma}_{\parallel n}$ is used in place of $\boldsymbol{\Sigma}_n$.
Parallel transport is necessary here as it can transfer information from one point to another by considering the rotation of the coordinate systems along the geodesic curve.
The transported covariance matrix can be computed as
\begin{equation}\label{parallel_cov}
\boldsymbol{\Sigma}_{\parallel n} = \sum_{j=1}^{d}\Gamma_{\boldsymbol{\mu}_n \to \boldsymbol{\mu}}(\mathfrak{u}_j)\Gamma_{\boldsymbol{\mu}_n \to \boldsymbol{\mu}}(\mathfrak{u}_j)^\top, 
\end{equation}
where $\mathfrak{u}_j$ is obtained through an eigendecomposition:
\begin{equation}\label{eigendecomp}
\boldsymbol{\Sigma}_n = \sum_{j=1}^{d}\mathfrak{u}_j\mathfrak{u}_j^\top.
\end{equation}

Similarly to the Euclidean setting, manifold-adaptive behavior, such as passing through some desired via-point, is realized by modifying the weight $\boldsymbol{\alpha}$ according to \eqref{weightalpha}.

Table \ref{T1} provides the manifold operations required for two problems of interest considered in this paper: The sphere $\mathbb{S}^2(r)$ with radius $r$ and the circular generalized cylinder $\mathbb{R}^2 \times \mathbb{S}$.
An algorithm for imitation with manifold-valued output is summarized in Algorithm \ref{manifold_alg}.

\section{Experimental Evaluations}\label{eval}
In this section, we evaluate the effectiveness of our proposed approach with both simulations and real experiments. 
Specifically, we conduct simulations on trajectory reproductions in Section \ref{trajrepro}, where the resulting imitation fidelity is reported.
We then test the performance of our algorithm in the case of adaptive behavior including both position and velocity via-points in Section \ref{trajadapt}.
For manifold trajectory learning, we study the problem of reproducing and adapting trajectory on a sphere and a generalized cylinder as in Section \ref{trajman}.
Finally, in Section \ref{realexp} we report the experimental results on a real KUKA robot arm.
A video of the real-world experiments is available in the supplementary material.

\begin{figure*}[t]
	\centering
	\begin{subfigure}[b]{1\textwidth}     
		\centering       
		\includegraphics[width=1\textwidth]{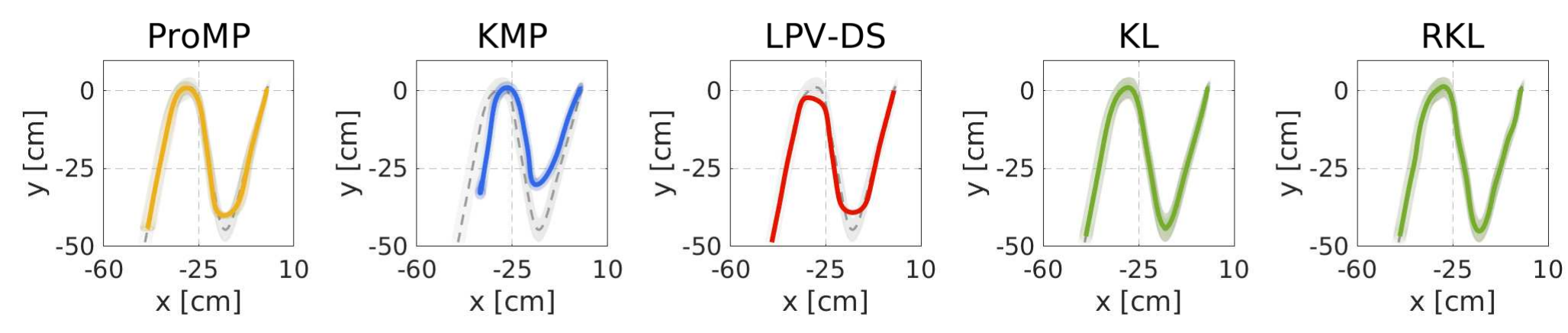}
		\label{fig:SRN}
	\end{subfigure}%
	\newline
	\begin{subfigure}[b]{1\textwidth}
		\centering
		\includegraphics[width=1\textwidth]{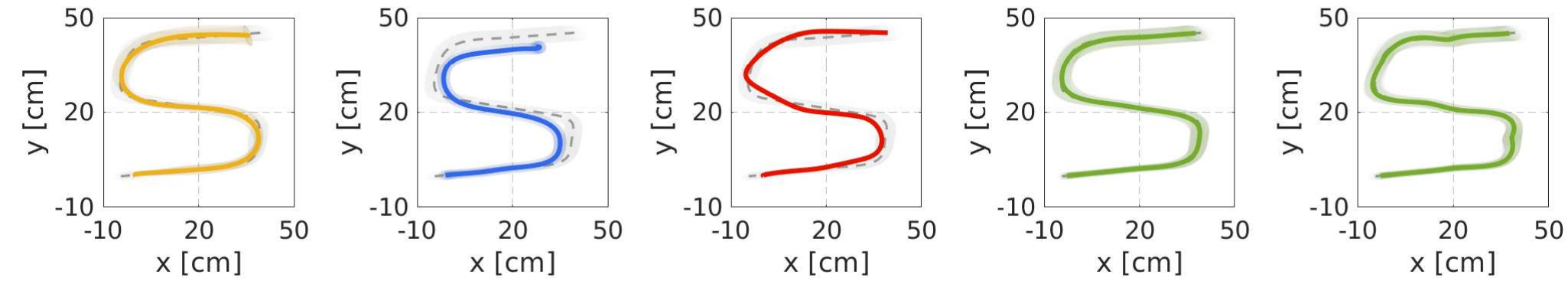}
		\label{fig:D-Imager}
	\end{subfigure}
	\begin{subfigure}[b]{1\textwidth}     
		\centering       
		\includegraphics[width=1\textwidth]{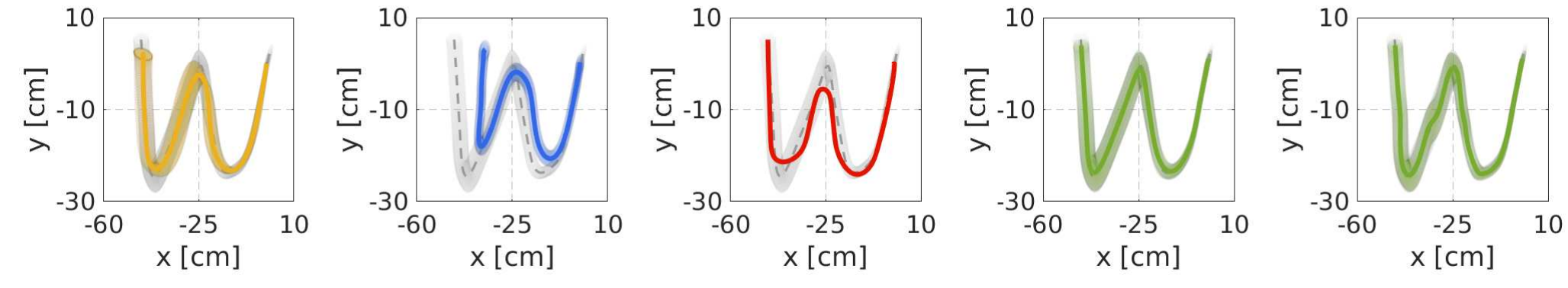}
		\label{fig:SRW}
	\end{subfigure}%
	\newline
	\begin{subfigure}[b]{1\textwidth}     
		\centering       
		\includegraphics[width=1\textwidth]{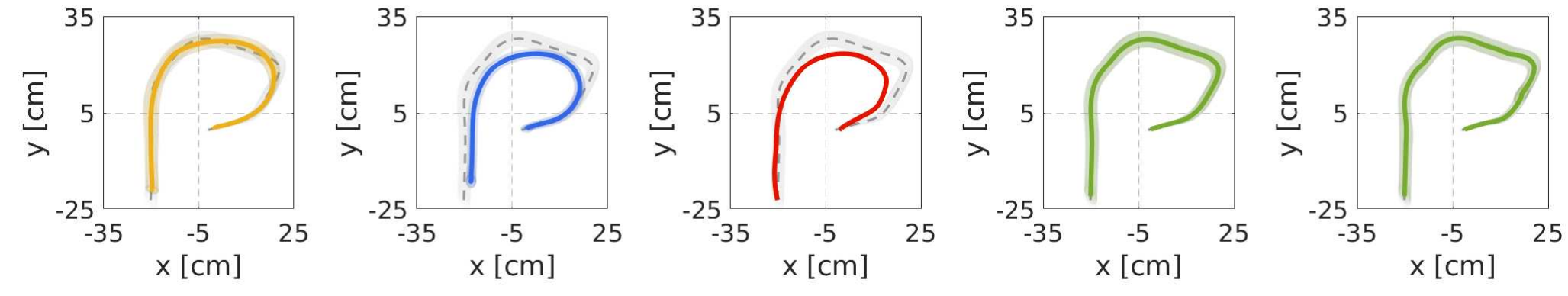}
		\label{fig:SRP}
	\end{subfigure}%
	\caption{Imitation fidelity comparison among ProMP (yellow), KMP (blue), LPV-DS (red) and our approach (green, with the fourth column KL imitation mode and the fifth column reverse KL imitation mode) for reproducing the trajectories of different letters. Solid curves and shallow areas are used to represent trajectory mean and standard deviation, respectively. The reference baseline used here is the probabilistic trajectory retrieved by GMR (gray).}\label{fig:TOF}
	\label{posrepro}
\end{figure*}

\subsection{Trajectory Reproduction}\label{trajrepro}
We first evaluate the effectiveness of our algorithm for trajectory reproduction.
Our goal is to compare with state-of-the-art algorithms in terms of imitation fidelity by learning the trajectories of four illustrative letters, namely 'N', 'S', 'W', and 'P', with time as the input. 
It should be noted that trajectory covariance learning is usually not considered by autonomous approaches like LPV-DS as they focus more on global stability and robustness to disturbances.

The collected data from multiple demonstrations for the letter trajectories are first processed with Gaussian mixture models (GMM). 
Subsequently, a probabilistic reference trajectory is extracted by Gaussian mixture regression (GMR) \citep{billard2008robot}, serving as the baseline for different imitation algorithms to compare against.

The learning results are shown in Figure \ref{posrepro}.
In the same row, the same multiple demonstrations for one letter are fed to different imitation learning algorithms in order to learn the mean and the covariance of a probabilistic trajectory. 
The first column in yellow represents ProMP.
ProMP is based on a parametric method, so a set of basis functions is needed to fit the demonstration trajectories.
We employ 30 Gaussian radial basis functions (RBF) for learning reference trajectories. 
The second column in blue represents KMP, whose hyperparameters, such as the regularization coefficients, are selected according to \citet{huang2019kernelized}.
The third column in red denotes LPV-DS.
In order for LPV-DS to generate a trajectory, an initial point needs to be manually selected.
Here, we set it to coincide with the reference trajectory's initial point.  
The last two columns in green represent our approach. 
Wherein, the fourth column denotes the KL divergence imitation mode and the fifth column denotes the reverse KL divergence imitation mode. 
To apply our algorithm, we choose the Gaussian kernel, which is defined by
\begin{equation}
k(\mathbf{x}_i, \mathbf{x}_j) = \exp(-\kappa\parallel\mathbf{x}_i-\mathbf{x}_j\parallel^2)
\end{equation}
where we set hyper-parameter $\kappa = 6$ when training the input-dependent weights $\boldsymbol{\alpha}$.

\begin{table*}[t]
	\caption{Performance comparison of different imitation learning algorithms} 
	\setlength{\tabcolsep}{5.6pt} 
	\renewcommand{\arraystretch}{1.2} 
	\centering
	\begin{tabular}{l||ccc||ccc||ccc||ccc}
		\hline
		& \multicolumn{3}{c||}{'N'} 
		& \multicolumn{3}{c||}{'S'}
		& \multicolumn{3}{c||}{'W'}
		& \multicolumn{3}{c}{'P'}\\
		\cline{2-13}
		& $C_m$ & $C_{cov}$ & Time ($\mathrm{s}$)& $C_m$ & $C_{cov}$ & Time ($\mathrm{s}$)& $C_m$ & $C_{cov}$ & Time ($\mathrm{s}$)& $C_m$ & $C_{cov}$ & Time ($\mathrm{s}$)\\
		\hline
		ProMP & 17.5 & 16.5 &0.01 & 15.8 &19.4 &0.02 &14.2 &17.7 &0.02 &14.6 &14.1 &0.01\\ \hline
		KMP & 40.3 & 19.3 & 1.84  &30.6 & 17.5  &1.83 &35.2 &18.4 &1.88 &29.2 &18.8 &2.10\\
		\hline
		LPV-DS & 60.8 & N/A &12.9   & 65.3& N/A &11.1 &61.0 &N/A  &13.5 &56.9 &N/A &9.5\\
		\hline
		Ours (KL) & 7.4 & 5.7 & 0.11  & 8.9&6.5 &0.09 &7.8 &6.0 &0.11 &5.9 &5.4 &0.09\\
		\hline
		Ours (RKL) & 7.3 & 6.4 &1.47  &9.3 &5.7 &1.51 &8.9 &6.2 &1.55 &5.0 &5.1 &1.52\\
		\hline
	\end{tabular}
	\label{comptable}
\end{table*}

In general, all imitation learning algorithms can correctly reproduce the originally demonstrated trajectories.
Both learned trajectory mean and covariance coincide with the reference probabilistic trajectory to some extent.
In Figure \ref{posrepro}, it can be observed that our proposed methods most closely reproduce the original demonstrations.
Especially, when it comes to abrupt turns, as in the case of the letter 'W', our algorithm can still imitate the demonstrated trajectory closely, while other methods exhibit a larger mismatch.

To quantitatively compare the reproduction performances of each algorithm, we report the cumulative error for trajectory mean and covariance, as well as the training time.
For evaluating the trajectory mean imitation, we choose the Root Mean Square Error (RMSE) between the estimated value and the demonstrated one: $C_{m} = \sqrt{ \sum_{n=1}^{N}\parallel\widehat{\mathbf{s}}_m(\mathbf{x}_n)- \boldsymbol{\mu}_n\parallel^2}$.   
Likewise, for the evaluation of trajectory covariance imitation, we choose the error as $C_{cov} = \sqrt{\sum_{n=1}^{N} \parallel\log(\boldsymbol{\Sigma}_n^{-\frac{1}{2}}\widehat{\mathbf{s}}_c(\mathbf{x}_n)\boldsymbol{\Sigma}_n^{-\frac{1}{2}})\parallel^2_{\mathtt{F}}}$, where the notation of geodesic distance of positive definite matrices is utilized.   
Concerning computational complexity, ProMP is expected to be more efficient since it grows as $\mathcal{O}(m^3d^3)$, where $m$ is the dimension of the basis function and $d$ is the dimension of output value.
KMP and our algorithm both have a $\mathcal{O}(n^3)$ time complexity, where $n$ denotes the number of trajectory data points.
LPV-DS is expected to be the most computationally expensive since a non-linear constrained optimization problem has to be solved.
The obtained numerical results are summarized in Table \ref{comptable}, where training time is averaged over five trials.
To conclude, our algorithm shows significant performance improvements with respect to the selected state-of-the-art methods on a diverse set of target trajectories, in terms of both trajectory mean and covariance imitation quality.

\begin{figure}[t]
	\centering       
	\includegraphics[width=0.96\columnwidth]{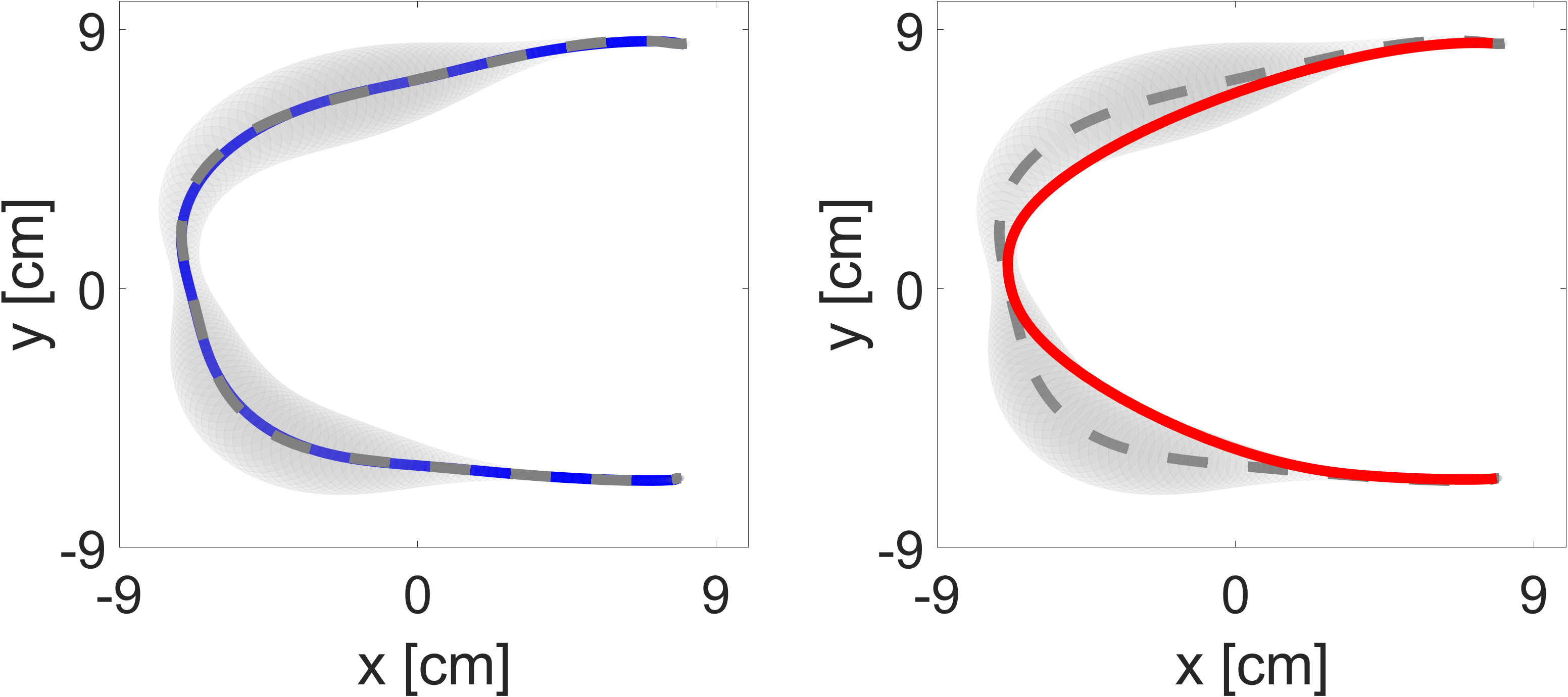}
	\caption{Comparison of imitation performances by the KL (blue) and RKL (red) imitation modes, where the dashed lines represent the reference mean and the shallow areas represent the trajectory covariance.}\label{fig:KLvsRKL}
\end{figure}

In addition, we would like to further clarify the difference in the imitation behaviors exhibited by the KL and RKL imitation modes.
In particular, we present the individual performance in terms of mean prediction given a C-shape probabilistic trajectory, as shown in Figure \ref{fig:KLvsRKL}.
It can be observed that when imitating in the KL mode, the reproduced trajectory closely follows the demonstrated trajectory mean as it is only dependent on the demonstrated trajectory mean. 
On the contrary, the reproduced trajectory using the RKL mode is not only dependent on the demonstrated trajectory mean but also on the demonstrated trajectory covariance, as evidenced by the larger deviation in the region with larger covariance. 
Different imitation modes will allow for more flexibility in probabilistic imitation learning.
For example, trajectory covariance can be exploited to prioritize trajectory mean prediction to balance task importance.

\subsection{Trajectory Adaptation}\label{trajadapt}
In this experiment, we study the problem of trajectory adaptation considering two scenarios: Position only and both position and velocity.
In position adaptation, three cases are studied, namely start-point, mid-way-point, and end-point adaptation.
Similarly to Section \ref{trajrepro}, ProMP and KMP are employed for comparison purposes.
LPV-DS is not considered here as it does not provide a straightforward manner for trajectory adaptation. 
We use the KL divergence imitation mode for our approach.
 
\begin{figure}[t]
	\centering
	\begin{subfigure}[b]{0.46\textwidth}
		\centering
		\includegraphics[width=1\textwidth]{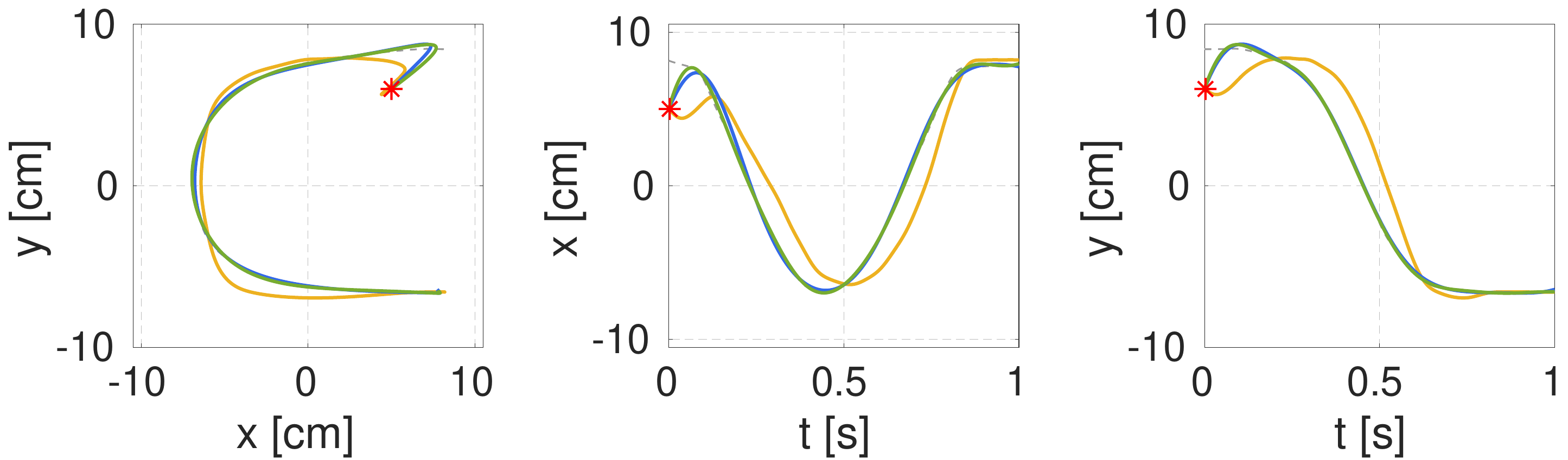}
		\caption{start-point adaptation}
		\label{fig:STARTPOINT}
	\end{subfigure}
	\begin{subfigure}[b]{0.46\textwidth}     
		\centering       
		\includegraphics[width=1\textwidth]{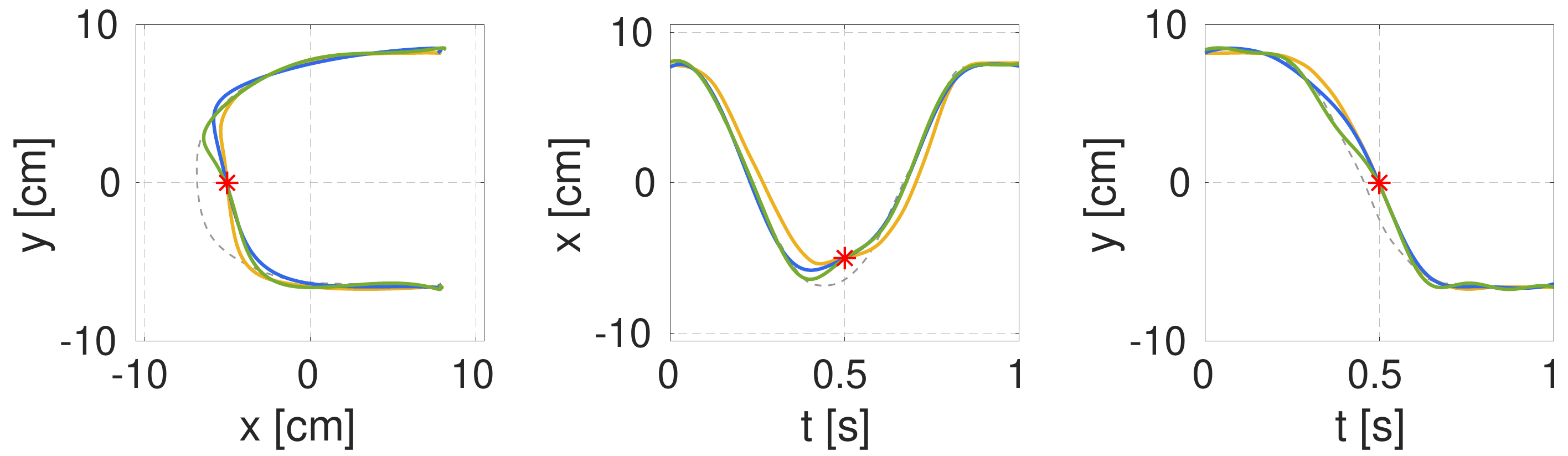}
		\subcaption{mid-way-point adaptation}
		\label{fig:SRMIDWAY}
	\end{subfigure}%
	\newline
	\begin{subfigure}[b]{0.46\textwidth}     
		\centering       
		\includegraphics[width=1\textwidth]{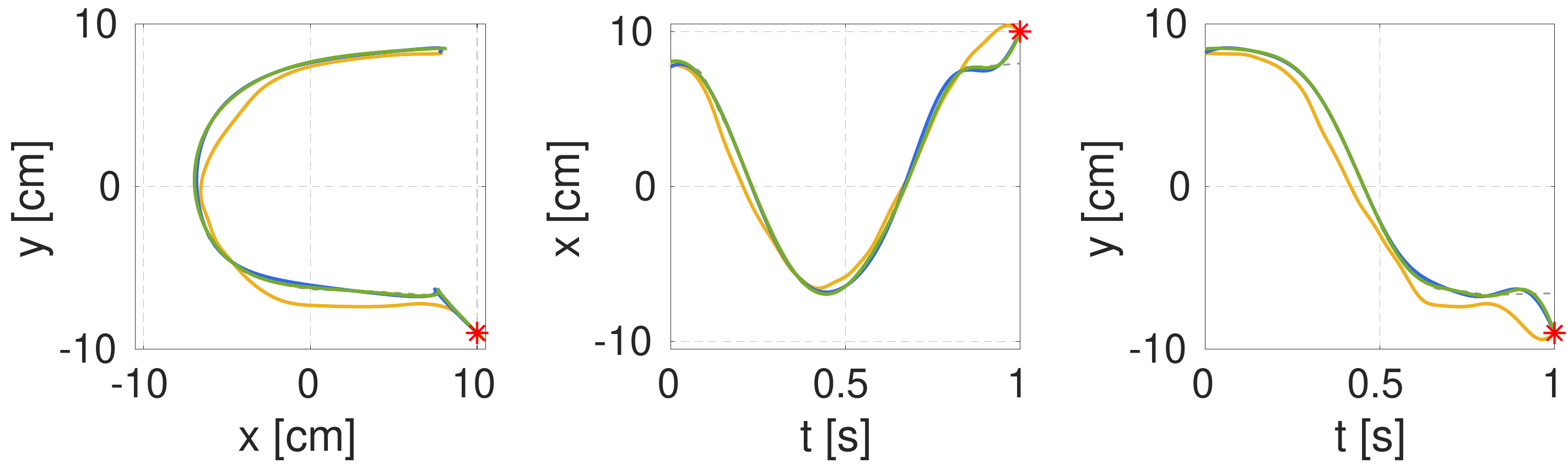}
		\subcaption{end-point adaptation}
		\label{fig:ENDPOINT}
	\end{subfigure}%
	\caption{Comparison of position adaptation among ProMP  (yellow), KMP (blue), and our approach (green) for a C-shape trajectory with time as the input. The desired points to go through are marked with red stars.}\label{fig:posadapt}
\end{figure}

For position adaptation, we consider modulating a C-shape trajectory to go through variously positioned via-points with time as the input.
We first evaluate start-point adaptation.
We set a desired point at $[5 \; 6]^\top \mathrm{cm}$ at time step $t = 0\, \mathrm{s}$.
Next, we move the desired point to $[-5 \; 0]^\top \mathrm{cm}$ at time step $t = 0.5\, \mathrm{s}$ to study mid-way-point adaptation.  
Finally, the issue of end-point adaptation is studied by setting the desired point at $[10 \; -9]^\top \mathrm{cm}$ at time step $t = 1.0\, \mathrm{s}$.
Different trajectory position adaptation cases are illustrated in Figure \ref{fig:posadapt}.
It shall be observed that all algorithms are capable of respecting the additional via-point constraint with high accuracy while the trajectory generated by our algorithm tends to converge to the original demonstrated trajectory closer compared with other methods.

\begin{figure}[t]
	\centering       
	\includegraphics[width=0.46\textwidth]{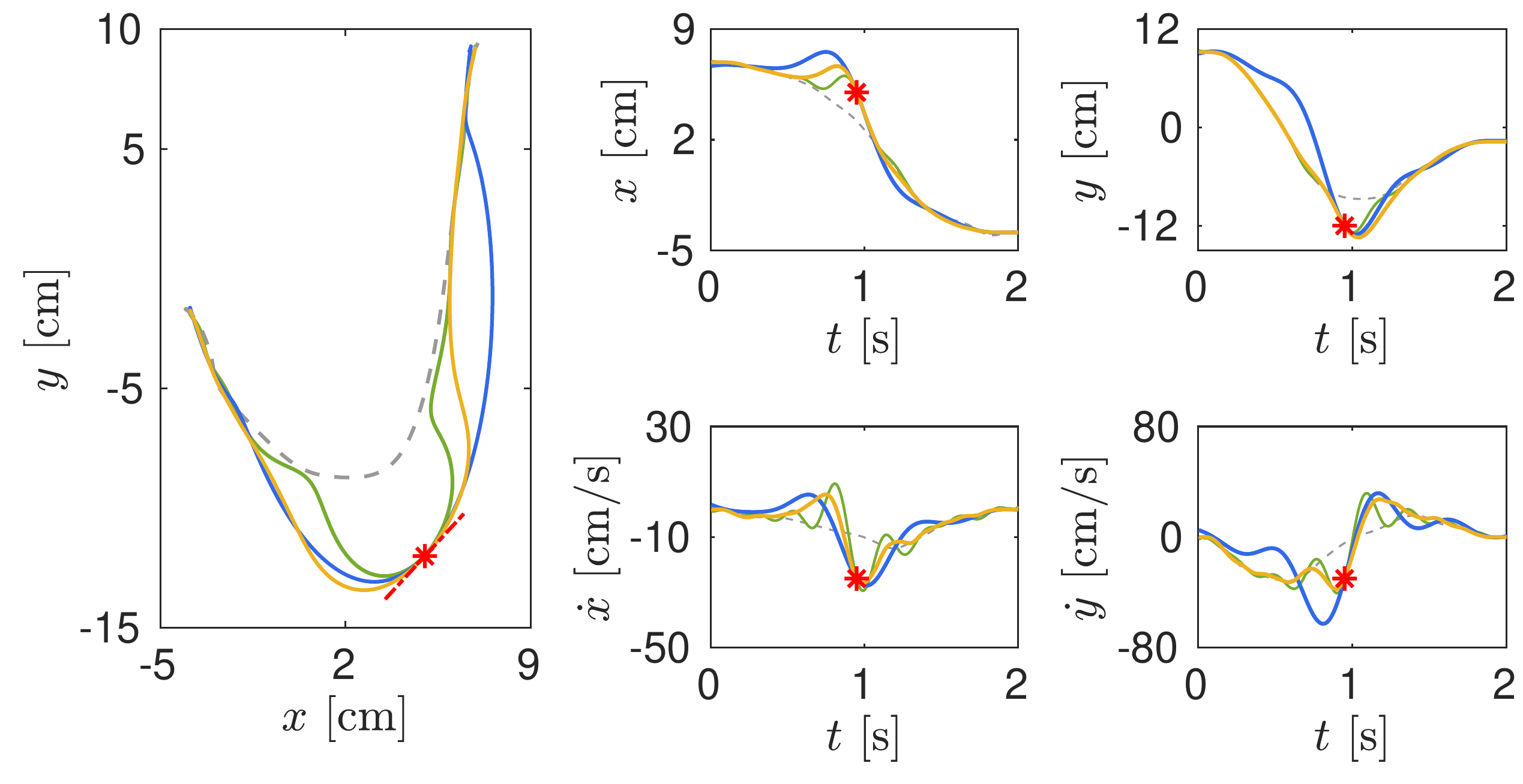}
	\caption{Comparison of both position and velocity adaptation among ProMP  (yellow), KMP (blue), and our approach (green) for a J-shape trajectory (dashed gray) with time as the input. The desired position points to go through are marked with red stars and the desired velocity is depicted with a dashed red line.}\label{fig:temporaladapt}
\end{figure}

\begin{figure*}[t]
	\centering
	\begin{subfigure}[b]{1\textwidth}
		\centering
		\includegraphics[width=1\textwidth]{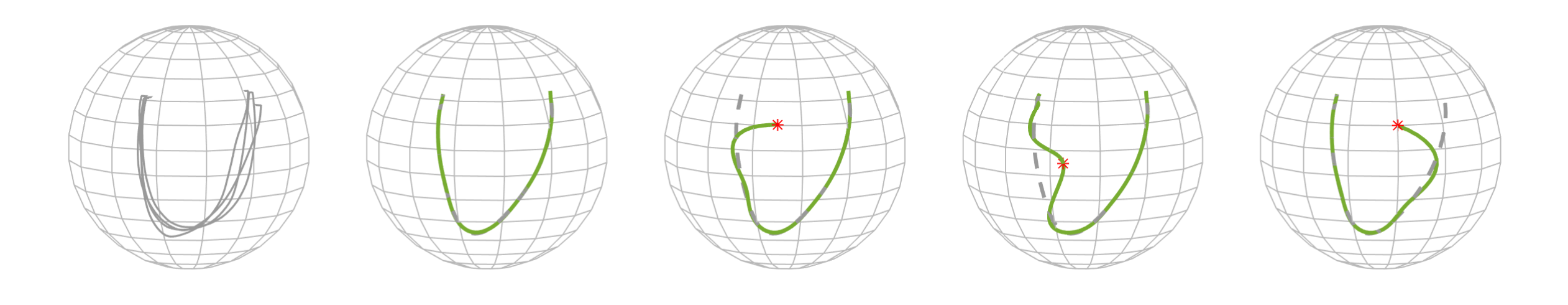}
		\caption{Trajectory imitation on a sphere}
		\label{fig:sphere}
	\end{subfigure}
	
	\begin{subfigure}[b]{1\textwidth}     
		\centering       
		\includegraphics[width=1\textwidth]{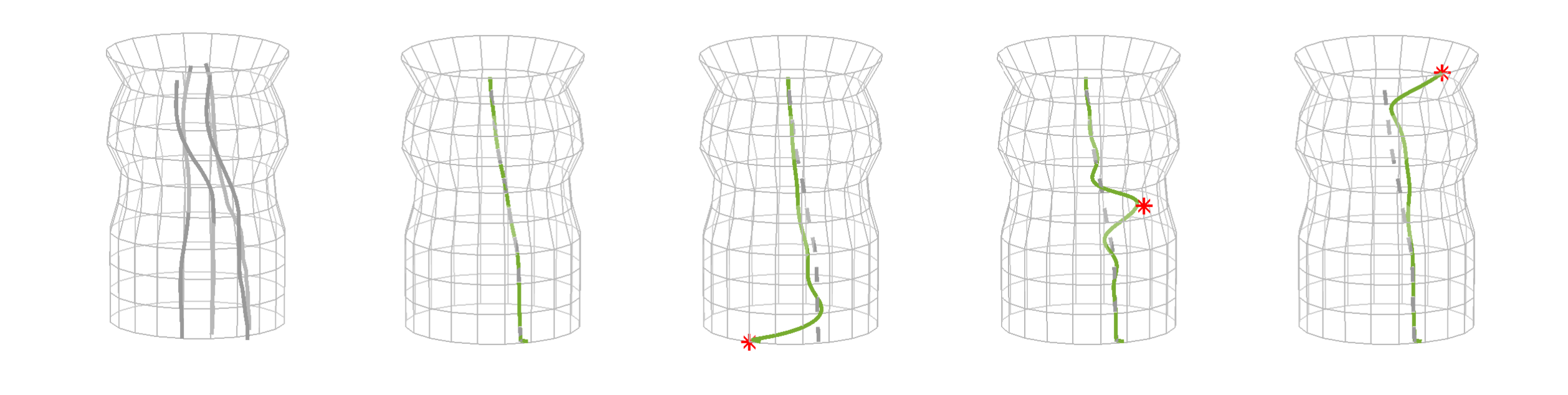}
		\subcaption{Trajectory imitation on a generalized cylinder}
		\label{fig:cylinder}
	\end{subfigure}
	\caption{Illustration of trajectory reproduction and adaptation on a manifold where multiple demonstrations are depicted in gray, trajectories generated by our algorithm are plotted in green and the desired via-point is marked with a red star.}\label{fig:trajman}
\end{figure*}

In the next experiment, we study the performance of adapting both the position and the velocity of an imitated trajectory.
We consider modulating a J-shape trajectory that has a total duration of $2\, \mathrm{s}$.
For ProMP, additional first-order derivatives of basis functions are needed to encode the dynamics of the motion.
By contrast, non-parametric methods like KMP and our approach do not require basis functions. 
To adapt the trajectory, at time step $t = 1 \, \mathrm{s}$ we set a desired location at $[5 \; -12]^\top \mathrm{cm}$ associated with a desired velocity $[-25 \; -30]^\top \mathrm{cm/s}$.
The obtained results are shown in Figure \ref{fig:temporaladapt}, where all the algorithms successfully generate trajectories that can pass through the desired position and desired velocity.
The adaptation errors at the via point are negligible.
It can be observed that the trajectory generated by our algorithm is affected the least by the additional adaptation requirements, while the other algorithms end up with larger imitation errors due to larger deviations.

\subsection{Manifold Trajectory Learning}\label{trajman}
In this experiment, we demonstrate a unique feature of our approach, namely, learning and adapting trajectories lying on manifolds.
To show the effectiveness of our approach, we study the problem of imitation on two common manifolds, namely a sphere ($\mathbb{S}^2$) and a circular generalized cylinder ($\mathbb{R}^2 \times \mathbb{S}$).  
The performance of both reproduction from multiple demonstrations as well as the adaptation towards a via point is evaluated.
Since adapting trajectories on manifolds has rarely been addressed before in robot imitation learning, we hereby only report the experimental results of our approach.

For learning on a sphere, we first draw four U-shaped trajectories on a sphere, each having time duration $t = 1 \, \mathrm{s}$, as shown in the first column of Figure \ref{fig:sphere}.
The radius of the sphere is $1 \, \mathrm{cm}$ and the center is located at the origin of the coordinate system.
The base point for conducting exponential and logarithmic mappings is chosen as $[0 \; 0 \; 1]^\top \mathrm{cm}$. 
Likewise, the collected demonstration data is processed using GMM and a reference probabilistic trajectory is subsequently extracted by GMR. 
As the collected data is defined on a Riemannian manifold, therefore, extended versions of GMM and GMR for Riemannian manifolds are employed for data processing \citep{zeestraten2017approach}. 

To show the adaptability on the manifold, a via-point at $[-0.199 \; -0.98 \; 0]^\top \mathrm{cm}$ is set on the surface of the sphere for the trajectory to pass through at $t_v = 0.35 \, \mathrm{s}$.
The second and third columns of Figure \ref{fig:sphere} show reproduction and adaptation trajectories from our algorithm, respectively. 
During each point prediction, the step size of Algorithm~\ref{manifold_alg} is set to $\eta = 0.01$. 
We can observe that our algorithm is capable of accurately meeting the requirement of via-point (marked with a red star) adaptation while preserving the shape of the original demonstrated trajectory.

The other experiment we carry out is trajectory imitation on a circular generalized cylinder, which has a smoothly varying circular cross-section.
We use the cylindrical coordinate system to express a data point $(r, z, \varphi)$, where $r$ is the radial distance from the $z$-axis, $z$ denotes height, and $\varphi$ is the azimuth. 
Since a data point lying on a generalized cylinder consists of a two-dimensional Euclidean component and a circular component, we formulate its Riemannian representation with the help of the Cartesian product: $\mathbb{R}^2\times\mathbb{S}^1$.
The center of the bottom of the cylinder is positioned at the origin of the cylindrical coordinate system and the cylinder center line is aligned with the $z$-axis.

We first draw four demonstrations along the cylinder's central line, each one of $1 \, \mathrm{cm}$ length and duration $t = 1 \, \mathrm{s}$, as shown in the first column of Figure \ref{fig:cylinder}.
The second column of Figure \ref{fig:cylinder} compares the reference retrieved by GMR and reproduction trajectory by our approach with step size again set to $\eta = 0.01$.
It can be observed that our algorithm is able to reproduce the reference trajectories with high imitation fidelity. 
At time step $t_v = 0.5 \, \mathrm{s}$, the trajectory is required to adapt towards a via-point located at $(1.51 \, \mathrm{cm}, 0.5 \, \mathrm{cm}, 0.863 \, \mathrm{rad})$.
The learning results are shown in the third column of Figure \ref{fig:cylinder}, where the adapted trajectory exactly passes through the via-point (marked with a red star).

\subsection{Polishing Task}\label{realexp}
The real-world experiment for evaluation of the proposed method is conducted via a polishing task.
The purpose of the polishing task is to teach a robot by kinesthetic guidance so that it learns how to polish on the surface of a sphere manifold. 
Once reproduction of the skill is achieved, the robot is later required to polish a user-defined via-point on the surface to exhibit adaptive behavior enabled by our approach. 
 
 \begin{figure}[t]
 	\centering       
 	\includegraphics[width=0.46\textwidth]{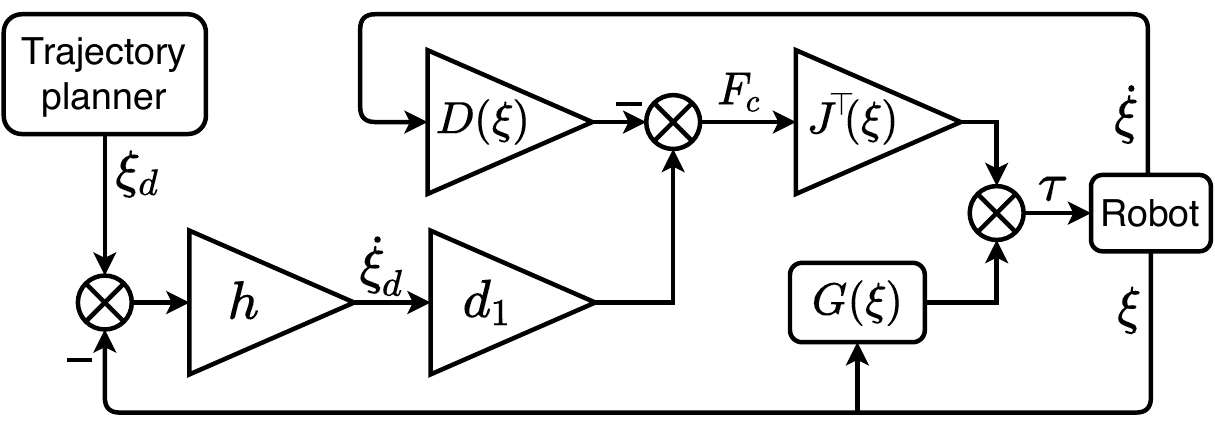}
 	\caption{Block illustration of the employed passive controller. The algorithm is composed of a velocity-based control law and a gravity-compensation term. }\label{fig:blkctrl}
 \end{figure}
 
 \subsubsection{Experimental setup}
The robot used for accomplishing the polishing task is the KUKA LWR IV+, a 7-DoF robotic arm. 
With the joint torque sensors mounted at the actuators, it can be torque-controlled and the torque commands are sent to KUKA using the Fast Research Interface (FRI) at $500 \,\mathrm{Hz}$. 
There is also a 6-axis force-torque sensor installed on the end-effector.
We choose to attach a 3D-printed finger tool to the robot as our end-effector.
In addition, a burst-resistant exercise ball is used as the sphere that the robot will polish on.

To transfer polishing skills from the human to the robot, we demonstrated the same task four times. 
During the demonstration, the robot is switched to the gravity compensation mode to make the robot light to be interacted with.
The Cartesian trajectory of the robot end-effector is recorded to train our algorithm.

\subsubsection{Robot control}
Tracking time-indexed position profiles usually poses a challenge for robots to achieve fast reactivity in response to external disturbances.
To this end, we employ a passive controller developed by \citet{kronander2015passive} to enhance the robot's robustness to large real-time disturbances.  
We consider the control of the robot's end effector in Cartesian space as in \citep{amanhoud2019dynamical}; the corresponding translational control force $\mathbf{F}_c$ takes the form
\begin{equation}\label{controller}
    \mathbf{F}_c = \mathbf{D}(\boldsymbol{\xi})(\dot{\boldsymbol{\xi}}_d - \dot{\boldsymbol{\xi}}) = d_1\dot{\boldsymbol{\xi}}_d - \mathbf{D}(\boldsymbol{\xi})\dot{\boldsymbol{\xi}},
\end{equation}
where $\boldsymbol{\xi}$ represents the end-effector position and $\mathbf{D}(\boldsymbol{\xi})$ is a state-varying damping matrix with the first eigen-vector $d_1 > 0$ aligned with the desired velocity $\dot{\boldsymbol{\xi}}_d$. 
For our time-invariant task representation, we design the desired velocity profile as
\begin{equation}\label{desvel}
    \dot{\boldsymbol{\xi}}_d = h(\boldsymbol{\xi}_d - \boldsymbol{\xi}),
\end{equation}
where $h>0$ weighs the tracking deviation.

In our experiment, the passive controller is only applied to the translational direction.
For orientation control, a control moment is computed by setting the end-effector roughly pointing towards the center of the ball using spherical linear interpolation with a PD control law.
Finally, the control commands are computed by mapping the control wrench, consisting of control moment and force, to joint torques using the corresponding robot Jacobian matrix.
Figure \ref{fig:blkctrl} presents a block representation of our proposed controller. 

\subsubsection{Results and analysis}
\begin{figure}[t]
	\centering       
	\includegraphics[width=0.49\textwidth]{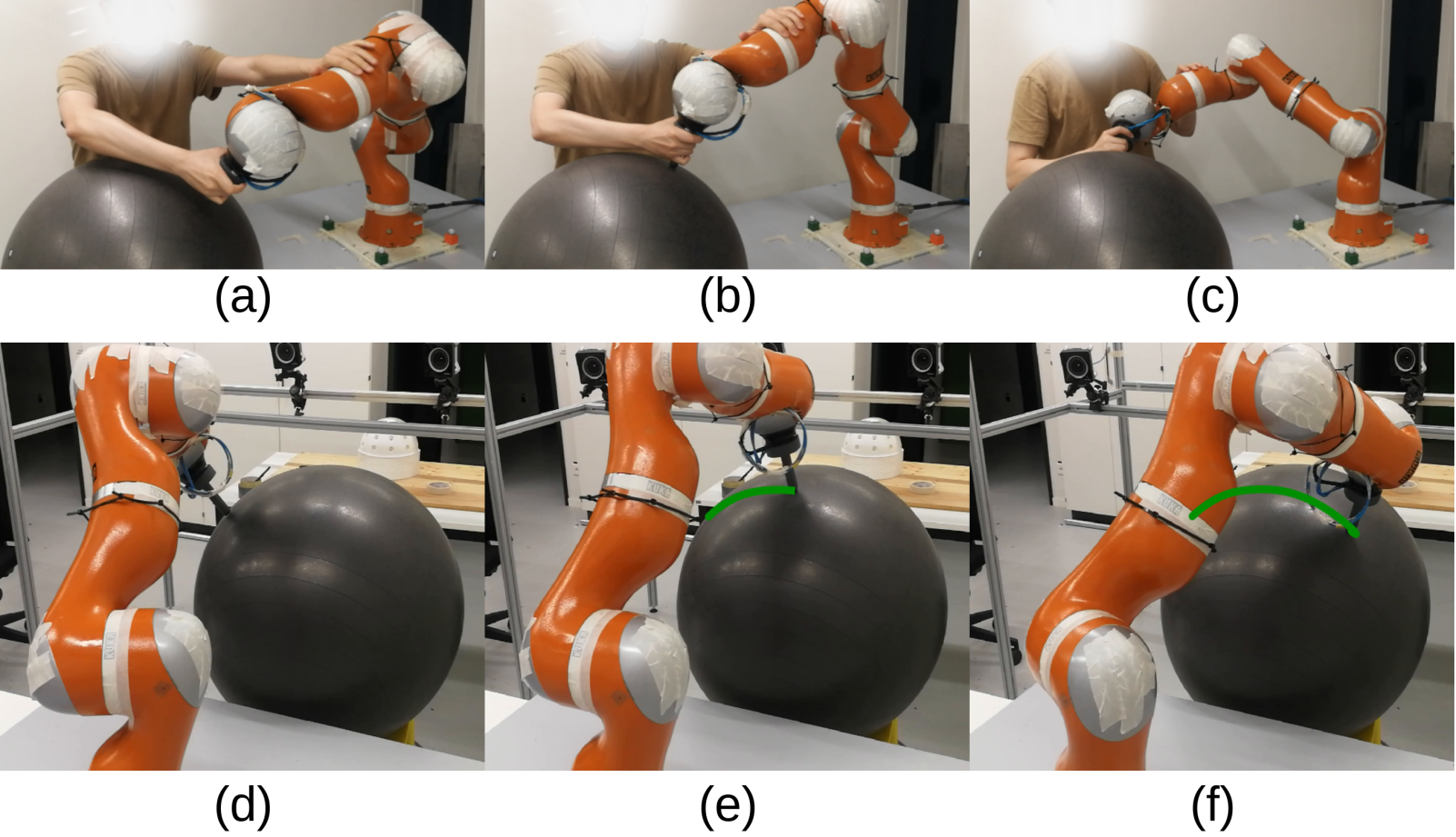}
	\caption{Snapshots of demonstration (\textit{top row}) and reproduction (\textit{bottom row}) of the polishing task with KUKA LWR IV+ robot arm.}\label{fig:demo_repro}
\end{figure}

An illustration of the kinesthetic demonstration procedure as well as skill reproduction is shown in Figure \ref{fig:demo_repro}.
A teacher demonstrates the polishing task multiple times to the robot on a sphere whose radius is $0.3 \mathrm{m}$ and the center position is identified as $[-0.7631 \; -0.0271 \; 0.0698]^\top \mathrm{m}$ in a least-squares sense.
We then evaluate the reproduction capabilities of our approach on the robot.
It shall be observed that the robot is able to reproduce the demonstrated task along the surface of the sphere.

\begin{figure*}[t]
	\centering
	\begin{subfigure}[b]{1\textwidth}     
		\centering       
		\includegraphics[width=1\textwidth]{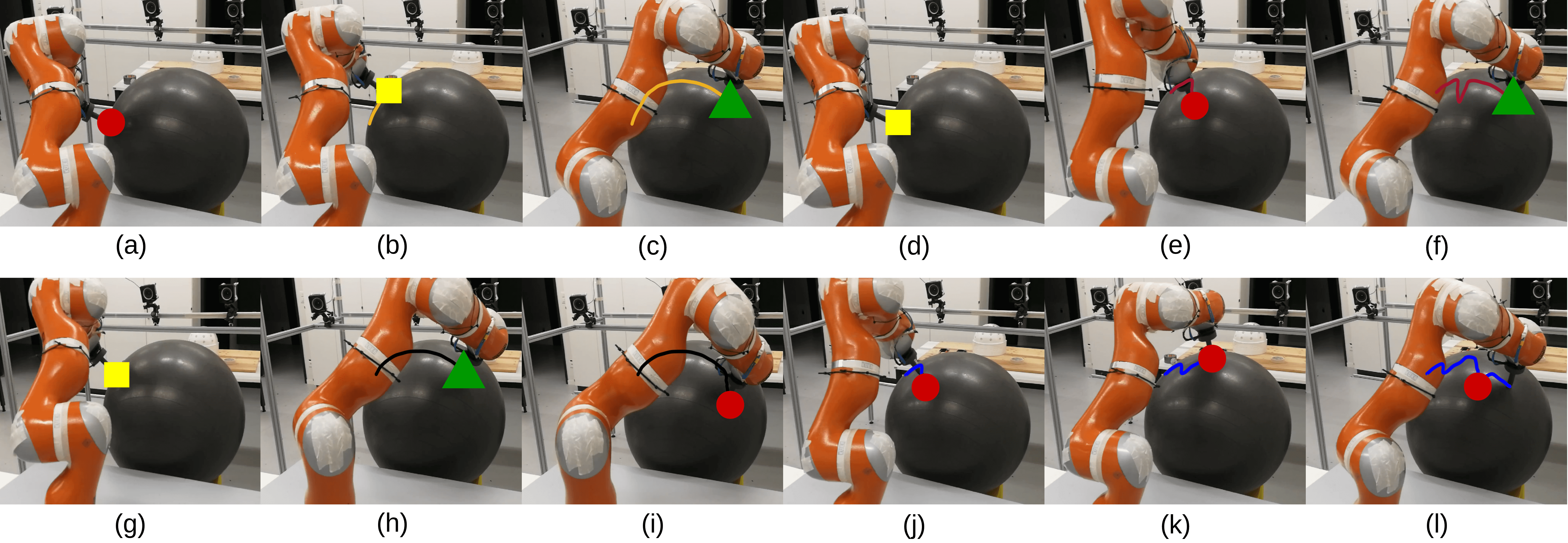}
	\end{subfigure}%
	\caption{Snapshots of the trajectory adaptation on a sphere with KUKA LWR IV+, in terms of a new desired start point((a)-(c)), a new mid-way point ((d)-(f)), a new desired endpoint ((g)-(i)) as well as multiple via-points((j)-(l)). The start points and the end points of the original demonstration trajectories and new desired points are marked with squares, triangles, and circles.}
	\label{realexpadapt}
\end{figure*}

\begin{figure}[t]
	\centering       
	\includegraphics[width=0.48\textwidth]{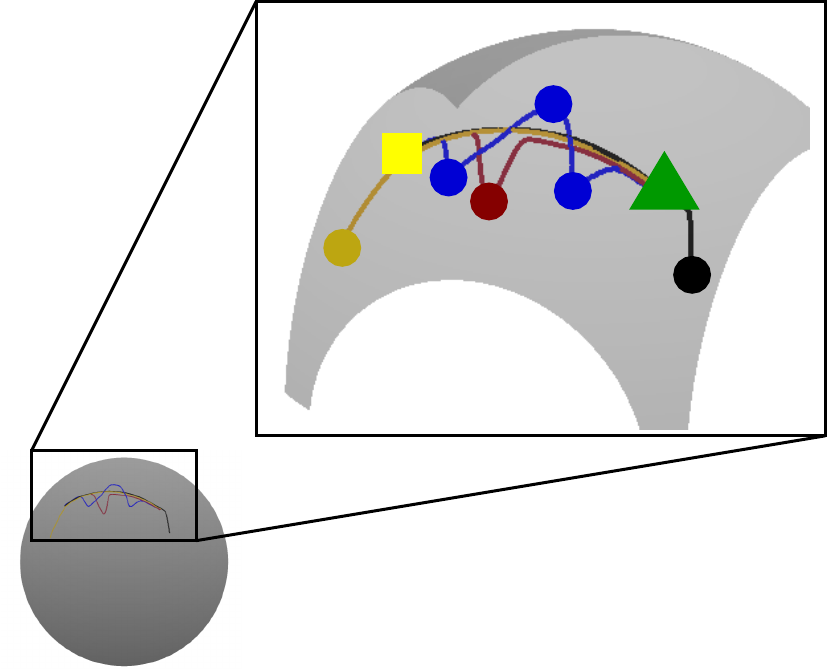}
	\caption{Plot of the robot end-effector adaptive trajectories in the task of polishing on the sphere, where the yellow, red, black, and blue curves represent start-point, mid-way-point, end-point, and multi-via-point adaptation, respectively.}\label{fig:realtraj}
\end{figure}

In another reproduction experiment, we apply external perturbations to the robot to test its working robustness.
By lifting the robot arm, the robot is still able to recover the originally planned trajectory.
The satisfactory compliant behavior makes it safe for the robot to interact with a human user.
To evaluate the adaptation capability of our approach, we set different via points for the robot end-effector to pass through.
Figure \ref{realexpadapt} shows that the robot end-effector movement is modulated to respect the constraint incurred by different additional desired via-points.
The real robot trajectories are plotted in Figure \ref{fig:realtraj}, where the relative position between the trajectories and the ball as well as the zoomed-in trajectories are provided for clarity.
Reproduction and adaptation performances are also reported in the supplementary video.

Though, in theory, the trajectory generated by our approach should exactly lie on the manifold, sometimes there is small interference or deviation observed between the end-effector and the ball.
This could be a result of the measuring error of the sphere information as well as the tracking error between the sent reference trajectory and the real robot end-effector position.
In the future, this issue could be resolved by introducing contact perception signals at the level of the control architecture design.

\section{Discussions}\label{discuss}
In this section, we first cover and compare with previous related work and analyze the connections (Section \ref{related}).
Then, we consider possible extensions to complement our proposed method and envisage future works (Section \ref{future}).  

\subsection{Related Work and Connections}\label{related}
The algorithmic foundations of robot imitation learning notably include regression techniques (see Table~\ref{comparison})
such as linear regression with basis functions (e.g., DMP and ProMP), kernel ridge regression (e.g., KMP), Gaussian mixture regression (e.g., SEDS), and Gaussian process regression (e.g., LGP and GPM), just to name a few. 
Although it is possible to bypass the regression step by calculating the arithmetic mean as done by Learning from demonstration by Averaging
Trajectories (LAT), the practical functionality of LAT can be severely limited due to the lack of trajectory adaptation.

Our method also follows the supervised learning paradigm. 
Remarkably, by leveraging structured prediction it achieves two main goals: The capability of prediction in structured output spaces and the flexibility of selecting different loss functions.

It is essential to investigate the most relevant types of output data for robot imitation learning since their correct processing is crucial for successful task execution.
Such output spaces usually possess some geometric structure \citep{Calinon20RAM}, e.g., when transferring skills from human experts to robots.
Compared with most movement primitives approaches that capture motion patterns by processing demonstrated data under the Euclidean metric, our approach distinguishes itself by the capability of prediction in structured output spaces such as Riemannian manifolds, which recently attracted growing interest in robotics  \citep{zeestraten2017approach, beik2021learning}.
Besides manifold-structured data, our framework preserves the potential to handle other output types (e.g., histograms, graphs, time series, etc.) thanks to the generality of the structured prediction paradigm.

It is also important to determine the similarity metric between the expert's and the learner's behavior.
Given the very same demonstrated dataset, different loss functions usually result in different imitation policies \citep{duan2020learning}.
Unlike conventional movement primitives that quadratically penalize deviation from the demonstrated trajectory, our approach permits a wide spectrum of loss functions thanks to the adopted implicit encoding framework for structured prediction.
To construct a proper loss function for probabilistic trajectory imitation, we take the novel point of view that imitation learning in sequential decision-making can be viewed as $f$-divergence minimization between learner and expert policy, as indicated by recent studies \citep{ke2019imitation, ghasemipour2019divergence}.

Being a common measure of the difference between two probability distributions, $f$-divergence also emerges as a popular tool in other robot learning settings.
In reinforcement learning algorithms such as Relative Entropy Policy Search (REPS) \citep{peters2010relative} and Trust Region Policy Optimization (TRPO) \citep{schulman2015trust}, a KL-divergence constraint is imposed between successive parameterized policies to prevent too large policy updates leading to unknown regions of the state space.  
Instead, in our approach the role of the $f$-divergence is to determine the coupling between the trajectory mean and covariance.
For example, when using the KL imitation mode for trajectory mean prediction, trajectory covariance is not involved.
By contrast, when using the RKL imitation mode for trajectory mean prediction, trajectory covariance is also employed.
This can provide an interface to incorporate the relative importance of demonstrations, i.e., small covariance represents high importance and will weigh more in trajectory generation \citep{huang2019kernelized}.
The different imitation modes as a result of different $f$-divergence functions thus enable multiple coupling modalities between trajectory mean and covariance in the case of probabilistic trajectory imitation.

Another related field is imitation for sequential decision-making, where a learner makes decisions by mimicking an expert given the demonstrated dataset of state-action pairs.
Notably, one of the fundamental issues in behavior cloning is the so-called compounding errors as formalized by DAgger~\citep{ross2011reduction}.
To alleviate the issue, DAgger iteratively appends training data to the dataset, yielding an interactive procedure for imitation.
Focusing on the form of the policy, implicit behavior cloning (IBC) casts imitation as a conditional energy-based modeling problem by using implicit models, leading to improvement in visuomotor tasks \citep{florence2022implicit}.
Aiming at providing a unifying framework for imitation, a game-theoretic perspective to imitation is proposed in~\citep{swamy2021moments} to minimize the worst-case divergence, and performance bounds for each imitation setting are analyzed.
Though sharing the same spirit of imitation, our approach directly concentrates on imitating robot trajectories, which gives rise to trajectory generation programmed by expert demonstrations.

Distilling the underlying reward, which is typically assumed to be linear in the features, is also a notable paradigm for imitation.
Based on the principle of maximum entropy, Maximum Entropy Inverse Reinforcement Learning (MaxEntIRL) recovers a reward by resolving the ambiguity in the matching of feature counts \citep{10.5555/1620270.1620297}.
Note that when using MaxEntIRL to generate robot trajectories a subsequent motion planning module is necessary \citep{9841604}. 
Geometric trajectory generation requires dedicated motion planning methods, such as \citep{bonallitrajectory} and \citep{kingston2019exploring},
which might complicate the procedure compared to our approach.

\subsection{Limitations and Future Work}\label{future}
For the sake of completeness, let us mention some current limitations in our proposed approach and potential future improvements.
As a memory-based approach, our algorithm may encounter difficulties when dealing with large-scale datasets (like other kernel methods).
Nevertheless, such a limitation may be alleviated by adopting sub-sampling techniques to reduce the kernel memory footprint \citep{rudi2015less}.
Besides, the requirement of performing Riemannian gradient descent for geometric trajectory imitation prevents the usage of our method on the fly.
Therefore, it may be beneficial to investigate more efficient Riemannian optimization strategies. 
In addition, when directly applying our method for making sequential decisions, it may suffer from compounding errors at test time.

For different imitation modes, we considered two representative $f$-divergence functions for the design of the loss function, namely the KL and the RKL divergences.  
With these two loss functions being SELF, it will be interesting to examine if other $f$-divergence functions also satisfy the SELF condition.
At first glance, it can be readily shown that the loss function is SELF when using the Jeffreys divergence given by $f(u) = (u-1)\log(u)$ (e.g., see \citep{pardo2018statistical}),
which is composed of the summation of the KL and the RKL divergences:
$D_{\mathtt{JD}}(\tilde{\mathbf{y}}_n, \tilde{\mathbf{y}}) = D_{\mathtt{KL}}(\tilde{\mathbf{y}}_n, \tilde{\mathbf{y}}) +
D_{\mathtt{RKL}}(\tilde{\mathbf{y}}_n, \tilde{\mathbf{y}}).$
For future work, we would like to investigate other types of $f$-divergence functions for the loss design.
In terms of applications, we conceive that our approach could be applied in a variety of robotic applications that require human-like
trajectory generation such as whole-body teleoperation for
humanoid robots \citep{10035484}.

\section{Conclusions}\label{concl}
In this paper, we present a novel robot imitation algorithm for probabilistic trajectory imitation. 
Specifically, an implicit embedding framework for structured prediction is employed, which endows our developed movement primitives with the capability of prediction with structured output space and the flexibility of choosing different loss functions.
Specifically, the loss functions used in the structured prediction are constructed by leveraging a novel point of view for imitation learning, i.e., minimization of the $f$-divergence between an expert's and a learner's policy.  
By choosing different types of $f$-divergences, we are able to learn demonstrated policies with different imitation modes.

We illustrate the effectiveness of our proposed approach by comparing it with state-of-the-art methods.
Our results demonstrate that our approach can increase imitation fidelity to a considerable extent in terms of both trajectory mean and covariance prediction while preserving merits such as multi-dimensional inputs and spatial and temporal trajectory modulations.   
Furthermore, our algorithm can be extended to learning and adapting trajectories on Riemannian manifolds, which distinguishes our approach from traditional methods. 

\begin{acks}
This research was supported by the Swiss National Science Foundation through the National Center of Competence in Research (NCCR) Robotics.
Part of this work has been carried out at the Machine Learning Genoa Center, Università di Genova, Italy. 
D. P. acknowledges the support by the An.Dy project, which has received funding from the European Union's Horizon 2020 Research and Innovation Programme under grant agreement No. 731540.
L. R. acknowledges the financial support of the European Research Council (grant SLING 819789), the Center for Brains, Minds and Machines, funded by NSF STC award CCF-1231216, the AFOSR projects FA9550-18-1-7009, FA9550-17-1-0390 and BAA-AFRL-AFOSR-2016-0007 (European Office of Aerospace Research and Development), the EU H2020-MSCA-RISE project NoMADS - DLV-777826, and the NVIDIA Corporation for the donation of a Titan Xp GPUs and a Tesla k40 GPU.
R. C. acknowledges the following: This study was carried out within the FAIR - Future Artificial Intelligence Research and received funding from the European Union Next-GenerationEU (PIANO NAZIONALE DI RIPRESA E RESILIENZA (PNRR) – MISSIONE 4 COMPONENTE 2, INVESTIMENTO 1.3 – D.D. 1555 11/10/2022, PE00000013). This manuscript reflects only the authors’ views and opinions, neither the European Union nor the European Commission can be considered responsible for them. 
\end{acks}

\appendix
\addcontentsline{toc}{section}{Appendices}
\renewcommand{\thesubsection}{\Alph{subsection}}
\section*{Appendix}\label{app}
\newcommand{\bigzero}{\mbox{\normalfont\large\bfseries 0}}
\subsection{Proof Sketch for SELF}\label{appSELF}
Here we provide sketch proof that the KL and the reverse KL divergence-based loss functions are SELF. 
Our proof is done by construction, i.e., we show that the loss functions can be expressed in the form of \eqref{defSELF}.
The central idea is to explicitly determine the feature map $\mathbf{c}$ and the linear operator $V$.
For convenience, we show that the mean and covariance cost functions can be constituted in a SELF fashion, which can readily lead to the loss functions being SELF.

\subsubsection{KL divergence}
Consider the following formulation for the cost function when making the mean prediction: 
\begin{align}
&(\boldsymbol{\mu}- \boldsymbol{\mu}_n)^\top\boldsymbol{\Sigma}^{-1}(\boldsymbol{\mu}- \boldsymbol{\mu}_n) \nonumber\\
=&\setlength\arraycolsep{1.9pt}{\underbrace{\left[
		\begin{array}{c}\boldsymbol{\mu}^\top\boldsymbol{\Sigma}^{-1}\boldsymbol{\mu}\\\boldsymbol{\mu}\\1
		\end{array}
		\right]}_{\textstyle\mathbf{c}(\boldsymbol{\mu})}}^{\top} 
\underbrace{\vphantom{\begin{array}{c}\boldsymbol{\mu}^\top\boldsymbol{\Sigma}^{-1}\boldsymbol{\mu}\\\boldsymbol{\mu}\\1
		\end{array}}\begin{array}{c}
	\mathtt{blkadiag}(1,\\-2\boldsymbol{\Sigma}^{-1},1)
	\end{array}}_{\textstyle V}
\underbrace{\left[
	\begin{array}{c}  \boldsymbol{\mu}_n^\top\boldsymbol{\Sigma}^{-1}\boldsymbol{\mu}_n \\\boldsymbol{\mu}_n\\1\end{array}
	\right]}_{\textstyle\mathbf{c}(\boldsymbol{\mu}_n)}\label{meanKLSELF}
\end{align}
where $\mathtt{blkadiag}(\cdot)$ denotes the block anti-diagonal matrix.

Consider the following formulation for the cost function when making covariance prediction: 
\begin{align}
&(\boldsymbol{\mu} \!-\! \boldsymbol{\mu}_n)^\top\boldsymbol{\Sigma}^{-1}(\boldsymbol{\mu}\!-\!\boldsymbol{\mu}_n) + \log|\boldsymbol{\Sigma}| + \mathtt{Tr}(\boldsymbol{\Sigma}^{-1}\boldsymbol{\Sigma}_n) \nonumber\\
=&\setlength{\arraycolsep}{1.9pt}{\underbrace{\left[\begin{array}{c}
		\boldsymbol{\mu}^\top\boldsymbol{\Sigma}^{-1}\boldsymbol{\mu}\\\boldsymbol{\Sigma}^{-1}\boldsymbol{\mu}\\\boldsymbol{\phi}_2(\boldsymbol{\Sigma}^{-\frac{1}{2}})\\ \log|\boldsymbol{\Sigma}|\\\mathtt{vec}(\boldsymbol{\Sigma})\\1\\\boldsymbol{\phi}_1(\boldsymbol{\mu})\\ -2\boldsymbol{\mu}\\1 \end{array}\right]}_{\textstyle\mathbf{c}(\boldsymbol{\Sigma})}}^\top \!\!\!\!\!
\underbrace{\vphantom{\left[\begin{array}{c}
		\boldsymbol{\mu}^\top\boldsymbol{\Sigma}^{-1}\boldsymbol{\mu}\\\boldsymbol{\Sigma}^{-1}\boldsymbol{\mu}\\\boldsymbol{\phi}_2(\boldsymbol{\Sigma}^{-\frac{1}{2}})\\ \log|\boldsymbol{\Sigma}|\\ \mathtt{vec}(\boldsymbol{\Sigma})\\1\\\boldsymbol{\phi}_1(\boldsymbol{\mu})\\ -2\boldsymbol{\mu}\\1 \end{array}\right]}
	\left[
	\begin{array}{c;{3.9pt/2pt}c}
	\raisebox{-0.6ex}[0ex][2ex]{\bigzero}\, & \begin{array}{c}
	\mathtt{blkadiag}\\(1,\mathbf{I}_{d}, \mathbf{I}_{q},
 \\1, \mathbf{0}_{d^2},\mathbf{I}_{d^2})
	\end{array} \\ \hdashline[3.9pt/2pt]
	\raisebox{-0.6ex}[1.9ex][1ex]{\bigzero}\, & \raisebox{-0.6ex}[1.9ex][1ex]{\bigzero} 
	\end{array}
	\right]}_{\textstyle V}
\underbrace{\vphantom{\left[\begin{array}{c}
		\boldsymbol{\mu}^\top\boldsymbol{\Sigma}^{-1}\boldsymbol{\mu}\\\boldsymbol{\Sigma}^{-1}\boldsymbol{\mu}\\\boldsymbol{\phi}_2(\boldsymbol{\Sigma}^{-\frac{1}{2}})\\ \log|\boldsymbol{\Sigma}|\\\mathtt{vec}(\boldsymbol{\Sigma})\\1\\\boldsymbol{\phi}_1(\boldsymbol{\mu})\\ -2\boldsymbol{\mu}\\1 \end{array}\right]}\!\!
	\left[\begin{array}{c}
	\boldsymbol{\mu}_n^\top\boldsymbol{\Sigma}_n^{-1}\boldsymbol{\mu}_n\\\boldsymbol{\Sigma}^{-1}_n\boldsymbol{\mu}_n\\\boldsymbol{\phi}_2(\boldsymbol{\Sigma}_n^{-\frac{1}{2}})\\ \log|\boldsymbol{\Sigma}_n|\\ \mathtt{vec}(\boldsymbol{\Sigma}_n)\\1\\\boldsymbol{\phi}_1(\boldsymbol{\mu}_n)\\ -2\boldsymbol{\mu}_n\\1 \end{array}\right]}_{\textstyle\mathbf{c}(\boldsymbol{\Sigma}_n)}
\end{align}
where $\mathbf{0}$ is the zeros matrix with proper dimension.
The features $\boldsymbol{\phi}_1(\cdot)$ and $\boldsymbol{\phi}_2(\cdot)$ are designed such that we have $\boldsymbol{\phi}_2(\boldsymbol{\Sigma}^{-\frac{1}{2}})^{\top}\boldsymbol{\phi}_1(\boldsymbol{\mu}_n) = \boldsymbol{\mu}_n\boldsymbol{\Sigma}^{-1}\boldsymbol{\mu}_n$. 
Also, we have
$\mathtt{vec}(\boldsymbol{\Sigma}) = 
\begin{bmatrix}   
\mathtt{vecr}(\boldsymbol{\Sigma})^\top & \mathtt{vecr}(\boldsymbol{\Sigma}^{-1})^\top & \mathtt{vecc}(\boldsymbol{\Sigma})^\top & \mathtt{vecc}(\boldsymbol{\Sigma}^{-1})^\top
\end{bmatrix}^\top$, 
where $\mathtt{vecr}(\cdot)$ and $\mathtt{vecc}(\cdot)$ denote the row- and column-major order vectorization operation for a matrix.

\subsubsection{Reverse KL divergence}
Similarly, the cost functions used in the reverse KL divergence can be expressed in the form of SELF.
For the cost function of mean prediction, consider the following formulation:
\begin{align}
&(\boldsymbol{\mu}- \boldsymbol{\mu}_n)^\top\boldsymbol{\Sigma}_n^{-1}(\boldsymbol{\mu}- \boldsymbol{\mu}_n) \nonumber\\
=& \setlength\arraycolsep{1.9pt}
{\underbrace{\left[
\begin{array}{c}
\boldsymbol{\mu}^\top\boldsymbol{\Sigma}^{-1}\boldsymbol{\mu}\\-2\boldsymbol{\mu}\\ \boldsymbol{\phi}_1(\boldsymbol{\mu})\\\boldsymbol{\phi}_2(\boldsymbol{\Sigma}^{-\frac{1}{2}})\\\boldsymbol{\Sigma}^{-1}\boldsymbol{\mu}\\1 
\end{array}
\right]}_{\textstyle\mathbf{c}(\boldsymbol{\mu})}}^\top \!\!\!\!\!
\underbrace{\vphantom{\left[
		\begin{array}{c}
		\boldsymbol{\mu}^\top\boldsymbol{\Sigma}^{-1}\boldsymbol{\mu}\\-2\boldsymbol{\mu}\\ \boldsymbol{\phi}_1(\boldsymbol{\mu})\\\boldsymbol{\phi}_2(\boldsymbol{\Sigma}^{-\frac{1}{2}})\\\boldsymbol{\Sigma}^{-1}\boldsymbol{\mu}\\1 
		\end{array}
		\right]}\begin{array}{c}
	\mathtt{blkadiag}\\(0,\mathbf{I}_{d}, \mathbf{I}_{q}\\\mathbf{0}_q, \mathbf{0}_d, 1)
	\end{array}}_{\textstyle V}
\underbrace{\vphantom{\left[
		\begin{array}{c}
		\boldsymbol{\mu}^\top\boldsymbol{\Sigma}^{-1}\boldsymbol{\mu}\\-2\boldsymbol{\mu}\\ \boldsymbol{\phi}_1(\boldsymbol{\mu})\\\boldsymbol{\phi}_2(\boldsymbol{\Sigma}^{-\frac{1}{2}})\\\boldsymbol{\Sigma}^{-1}\boldsymbol{\mu}\\1 
		\end{array}
		\right]}
	\left[
\begin{array}{c}
\boldsymbol{\mu}_n^\top\boldsymbol{\Sigma}_n^{-1}\boldsymbol{\mu}_n\\-2\boldsymbol{\mu}_n\\ \boldsymbol{\phi}_1(\boldsymbol{\mu}_n)\\\boldsymbol{\phi}_2(\boldsymbol{\Sigma}_n^{-\frac{1}{2}})\\ \boldsymbol{\Sigma}_n^{-1}\boldsymbol{\mu}_n\\1 
\end{array}
\right]}_{\textstyle\mathbf{c}(\boldsymbol{\mu}_n)}
\end{align}

For covariance prediction, consider the following formulation for the cost function:
\begin{align}
&-\log|\boldsymbol{\Sigma}| +
\mathtt{Tr}(\boldsymbol{\Sigma}_n^{-1}\boldsymbol{\Sigma}) \nonumber\\
=&\setlength\arraycolsep{1.9pt} 
{\underbrace{
\left[
\begin{array}{c} \log|\boldsymbol{\Sigma}| \\ \mathtt{vec}(\boldsymbol{\Sigma}) \\ 1 \end{array}\right]}_{\textstyle\mathbf{c}(\boldsymbol{\Sigma})}}^\top \!\!\!
\underbrace{\vphantom{\left[
		\begin{array}{c} \log|\boldsymbol{\Sigma}| \\ \mathtt{vec}(\boldsymbol{\Sigma}) \\ 1 \end{array}\right]}
\left[
\begin{array}{c;{3.9pt/2pt}c}
\raisebox{-0.6ex}[0ex][2ex]{\bigzero}\, & \begin{array}{c}
\mathtt{blkadiag}(\\-1,\mathbf{I}_{d^2})
\end{array} \\ \hdashline[3.9pt/2pt]
\raisebox{-0.6ex}[1.9ex][1ex]{\bigzero}\, & \raisebox{-0.6ex}[1.9ex][1ex]{\bigzero} 
\end{array}
\right]}_{\textstyle V}
\underbrace{\left[
\begin{array}{c} \log|\boldsymbol{\Sigma}_n| \\ 
\mathtt{vec}(\boldsymbol{\Sigma}_n)  \\
1 
\end{array}
\right]}_{\textstyle\mathbf{c}(\boldsymbol{\Sigma}_n)}
\end{align}

\subsection{Definitions for Riemannian manifold}\label{Rie}
\begin{figure}[t]
	\centering
	\begin{subfigure}[b]{0.529\textwidth}
		\centering
		\includegraphics[width=0.529\textwidth]{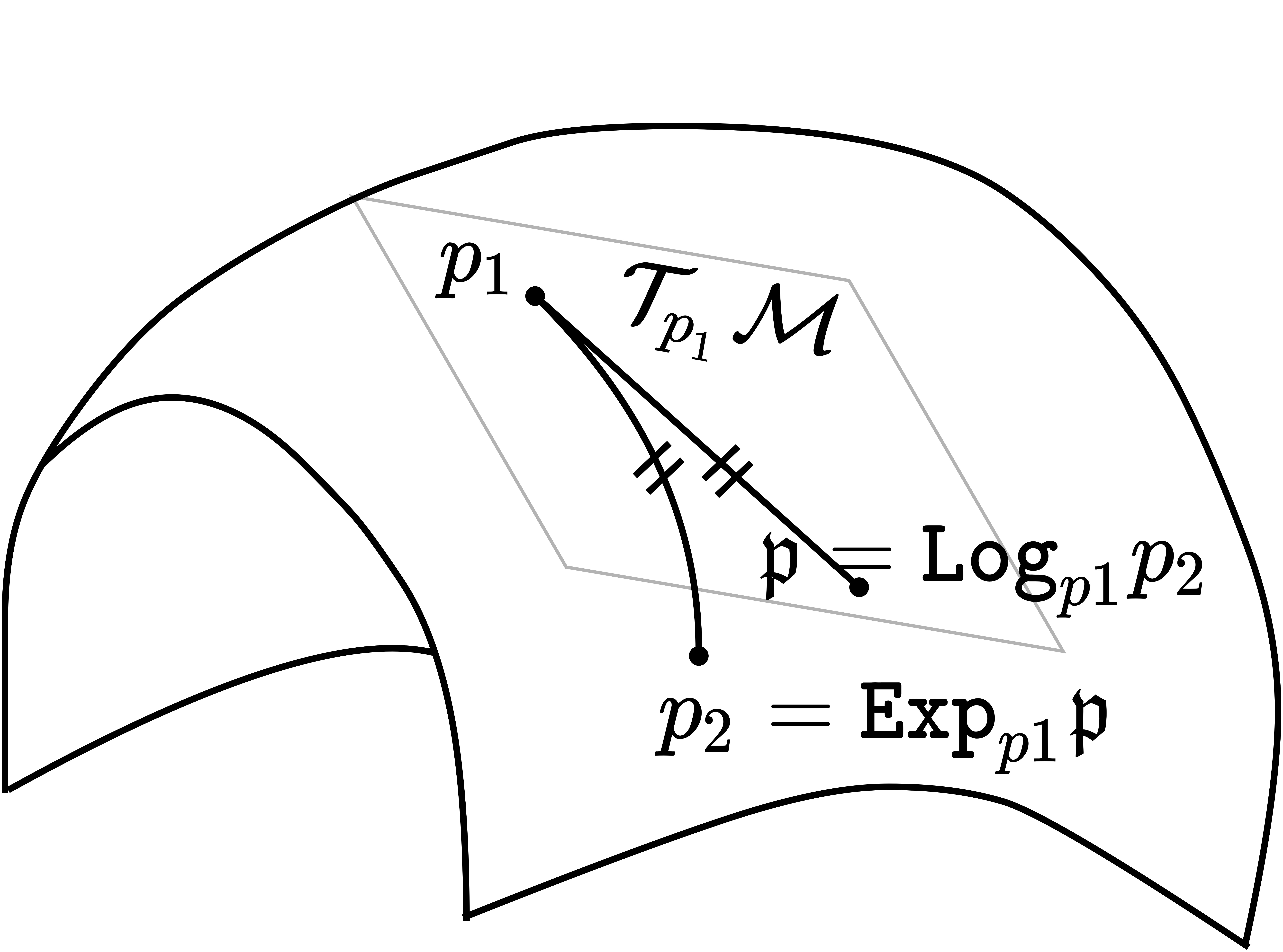}
		\caption{Manifold exponential and logarithmic mappings.}
		\label{mangeo}
	\end{subfigure}
	\begin{subfigure}[b]{0.6\textwidth}
		\centering
		\includegraphics[width=0.6\textwidth]{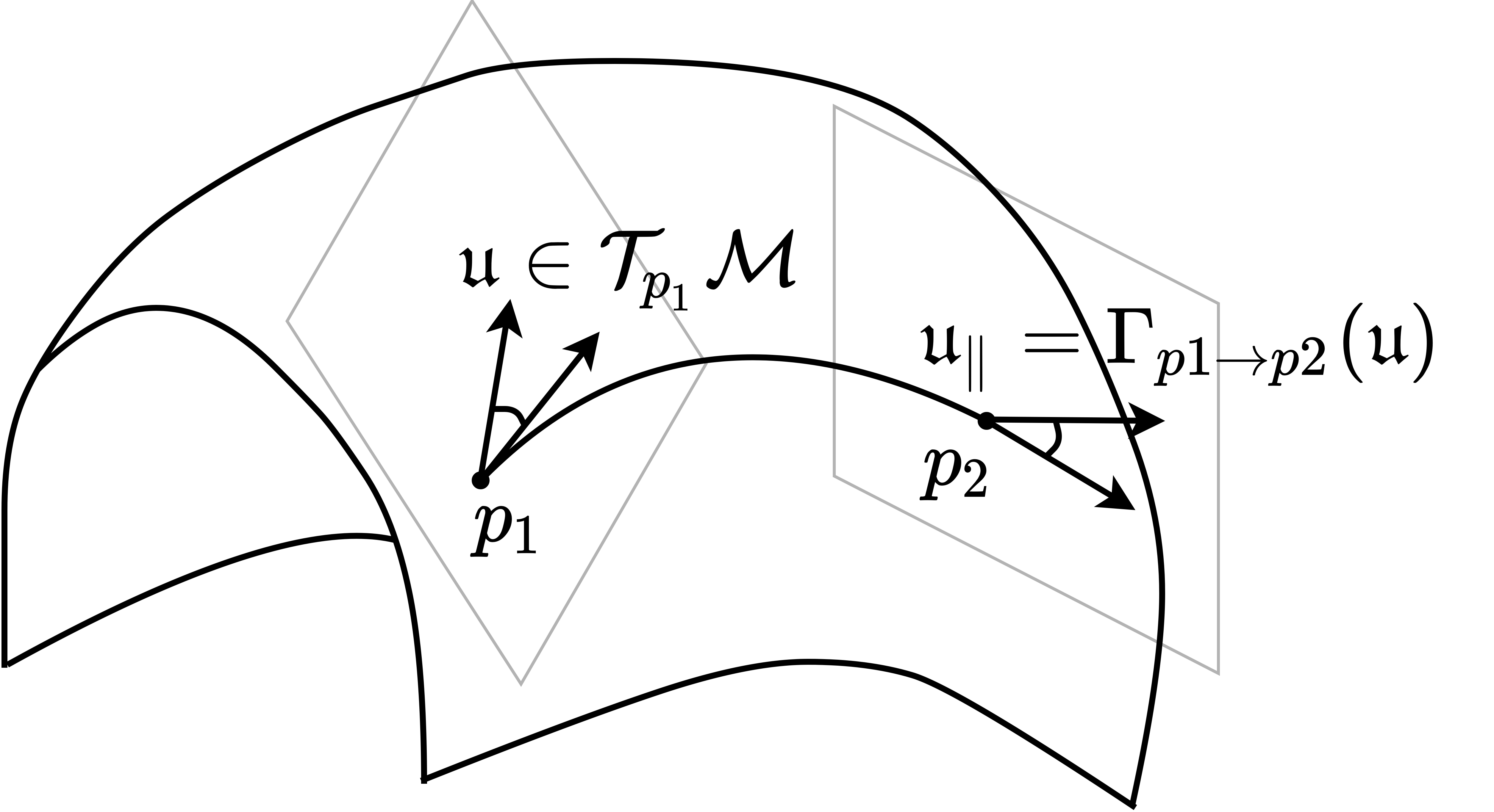}
		\caption{Parallel transport of vector.}
		\label{fig:mantransp}
	\end{subfigure}
	\caption{Illustration of basic Riemannian notions.}
	\label{fig:illuman}
\end{figure}

Here we provide basic notions and corresponding notations on Riemannian statistics that are used throughout this paper. 
We refer interested readers to \citet{absil2009optimization} for more details on manifolds.
Informally, a $d$-dimensional Riemannian manifold $(\mathcal{M}, g)$ is a topological space that locally behaves like the Euclidean space $\mathbb{R}^{d}$.
Every point $\mathbf{p} \in \mathcal{M}$ has a tangent space
$\mathcal{T}_{\mathbf{p}}\mathcal{M}$ where an inner product $g$ is defined.
When $\mathcal{M}$ is a \textit{submanifold} of $\mathbb{R}^{d+1}$, the inner product can inherit from the standard Euclidean inner product in a natural way.  
The minimum distance between two points $\mathbf{p}_1$ and $\mathbf{p}_2$ on a Riemannian manifold is called \textit{geodesic distance} $\mathtt{dist}(\mathbf{p}_1, \mathbf{p}_2)$, which generalizes the concept of straight lines in Euclidean spaces.
The \textit{exponential map} $\mathtt{Exp}_{\mathbf{p}}: \mathcal{T}_{\mathbf{p}}\mathcal{M} \rightarrow \mathcal{M}$ maps a point in the tangent space to the manifold with the distance and direction preserved.
A \textit{retraction} map $R_{\mathbf{p}}: \mathcal{T}_{\mathbf{p}}\mathcal{M} \rightarrow \mathcal{M}$ is a first order approximation of the exponential map.
The inverse mapping of the exponential map is called the \textit{logarithm map} $\mathtt{Log}_{\mathbf{p}}: \mathcal{M} \rightarrow \mathcal{T}_{\mathbf{p}}\mathcal{M}$.
The logarithmic map is defined except for the \textit{cut locus} of the base, which is the set of points that are connected by more than one geodesic curve with the base.
\textit{Parallel transport} $\Gamma_{\mathbf{p}_1 \to \mathbf{p}_2}(\mathfrak{u}): \mathcal{T}_{\mathbf{p}_1}\mathcal{M} \rightarrow \mathcal{T}_{\mathbf{p}_2}\mathcal{M}$ moves vectors between tangent spaces such that the angles between the vectors and the geodesic curve connecting the bases are conserved.
This operation is necessary to transport information available in one tangent space to another.
The \textit{Cartesian product} of two Riemannian manifolds $\mathcal{M}_1 \times \mathcal{M}_2$ is still a Riemannian manifold and the corresponding manifold operations mentioned above can be obtained by concatenating the individual operations.
A pictorial illustration is shown in Figure \ref{fig:illuman}.

\bibliographystyle{SageH}
\bibliography{bibiography}

\begin{thebibliography}{61}
\providecommand{\natexlab}[1]{#1}
\providecommand{\url}[1]{\texttt{#1}}
\providecommand{\urlprefix}{URL }
\expandafter\ifx\csname urlstyle\endcsname\relax
  \providecommand{\doi}[1]{DOI:\discretionary{}{}{}#1}\else
  \providecommand{\doi}{DOI:\discretionary{}{}{}\begingroup \urlstyle{rm}\Url}\fi

\bibitem[{Abbeel et~al.(2010)Abbeel, Coates and Ng}]{abbeel2010autonomous}
Abbeel P, Coates A and Ng AY (2010) Autonomous helicopter aerobatics through apprenticeship learning.
\newblock \emph{The International Journal of Robotics Research} 29(13): 1608--1639.

\bibitem[{Abbeel and Ng(2004)}]{abbeel2004apprenticeship}
Abbeel P and Ng AY (2004) Apprenticeship learning via inverse reinforcement learning.
\newblock In: \emph{Proceedings of the twenty-first international conference on Machine learning}. p.~1.

\bibitem[{Absil et~al.(2009)Absil, Mahony and Sepulchre}]{absil2009optimization}
Absil PA, Mahony R and Sepulchre R (2009) \emph{Optimization algorithms on matrix manifolds}.
\newblock Princeton University Press.

\bibitem[{Ahmadzadeh and Chernova(2018)}]{ahmadzadeh2018trajectory}
Ahmadzadeh SR and Chernova S (2018) Trajectory-based skill learning using generalized cylinders.
\newblock \emph{Frontiers in Robotics and AI} 5.

\bibitem[{Ajoudani et~al.(2018)Ajoudani, Fang, Tsagarakis and Bicchi}]{ajoudani2018reduced}
Ajoudani A, Fang C, Tsagarakis N and Bicchi A (2018) Reduced-complexity representation of the human arm active endpoint stiffness for supervisory control of remote manipulation.
\newblock \emph{The International Journal of Robotics Research} 37(1): 155--167.

\bibitem[{{\'A}lvarez et~al.(2012){\'A}lvarez, Rosasco and Lawrence}]{alvarez2012kernels}
{\'A}lvarez MA, Rosasco L and Lawrence ND (2012) Kernels for vector-valued functions: A review.
\newblock \emph{Foundations and Trends{\textregistered} in Machine Learning} 4(3): 195--266.

\bibitem[{Amanhoud et~al.(2019)Amanhoud, Khoramshahi and Billard}]{amanhoud2019dynamical}
Amanhoud W, Khoramshahi M and Billard A (2019) A dynamical system approach to motion and force generation in contact tasks.
\newblock Robotics: Science and Systems (RSS).

\bibitem[{Arduengo et~al.(2021)Arduengo, Colomé, Borràs, Sentis and Torras}]{9345363}
Arduengo M, Colomé A, Borràs J, Sentis L and Torras C (2021) Task-adaptive robot learning from demonstration with {G}aussian process models under replication.
\newblock \emph{IEEE Robotics and Automation Letters} 6(2): 966--973.
\newblock \doi{10.1109/LRA.2021.3056367}.

\bibitem[{Bahl et~al.(2020)Bahl, Mukadam, Gupta and Pathak}]{bahl2020neural}
Bahl S, Mukadam M, Gupta A and Pathak D (2020) Neural dynamic policies for end-to-end sensorimotor learning.
\newblock \emph{Advances in Neural Information Processing Systems} 33: 5058--5069.

\bibitem[{Beik-Mohammadi et~al.(2021)Beik-Mohammadi, Hauberg, Arvanitidis, Neumann and Rozo}]{beik2021learning}
Beik-Mohammadi H, Hauberg S, Arvanitidis G, Neumann G and Rozo L (2021) Learning {Riemannian} manifolds for geodesic motion skills.
\newblock In: \emph{Robotics: Science and Systems}.

\bibitem[{Billard et~al.(2008)Billard, Calinon, Dillmann and Schaal}]{billard2008robot}
Billard A, Calinon S, Dillmann R and Schaal S (2008) Robot programming by demonstration.
\newblock \emph{Springer handbook of robotics} : 1371--1394.

\bibitem[{Bonalli et~al.(2019)Bonalli, Bylard, Cauligi, Lew and Pavone}]{bonallitrajectory}
Bonalli R, Bylard A, Cauligi A, Lew T and Pavone M (2019) Trajectory optimization on manifolds: A theoretically-guaranteed embedded sequential convex programming approach.
\newblock In: \emph{Robotics: {Science} and {Systems}}.

\bibitem[{Calinon(2020)}]{Calinon20RAM}
Calinon S (2020) Gaussians on riemannian manifolds: Applications for robot learning and adaptive control.
\newblock \emph{IEEE Robotics \& Automation Magazine} 27(2): 33--45.

\bibitem[{Cheng et~al.(2018)Cheng, Yan, Wagener and Boots}]{cheng2018fast}
Cheng CA, Yan X, Wagener N and Boots B (2018) Fast policy learning through imitation and reinforcement.
\newblock In: \emph{Uncertainty in artificial intelligence}.

\bibitem[{Ciliberto et~al.(2016)Ciliberto, Rosasco and Rudi}]{ciliberto2016consistent}
Ciliberto C, Rosasco L and Rudi A (2016) A consistent regularization approach for structured prediction.
\newblock In: \emph{Advances in neural information processing systems}. pp. 4412--4420.

\bibitem[{Ciliberto et~al.(2020)Ciliberto, Rosasco and Rudi}]{ciliberto2020general}
Ciliberto C, Rosasco L and Rudi A (2020) A general framework for consistent structured prediction with implicit loss embeddings.
\newblock \emph{J. Mach. Learn. Res.} 21(98): 1--67.

\bibitem[{Darvish et~al.(2023)Darvish, Penco, Ramos, Cisneros, Pratt, Yoshida, Ivaldi and Pucci}]{10035484}
Darvish K, Penco L, Ramos J, Cisneros R, Pratt J, Yoshida E, Ivaldi S and Pucci D (2023) Teleoperation of humanoid robots: A survey.
\newblock \emph{IEEE Transactions on Robotics} : 1--22\doi{10.1109/TRO.2023.3236952}.

\bibitem[{Duan et~al.(2018)Duan, Camoriano, Ferigo, Calandriello, Rosasco and Pucci}]{duan2018constrained}
Duan A, Camoriano R, Ferigo D, Calandriello D, Rosasco L and Pucci D (2018) Constrained {DMP}s for feasible skill learning on humanoid robots.
\newblock In: \emph{2018 IEEE-RAS 18th International Conference on Humanoid Robots (Humanoids)}. IEEE, pp. 1--6.

\bibitem[{Duan et~al.(2019)Duan, Camoriano, Ferigo, Huang, Calandriello, Rosasco and Pucci}]{duanlearning}
Duan A, Camoriano R, Ferigo D, Huang Y, Calandriello D, Rosasco L and Pucci D (2019) Learning to sequence multiple tasks with competing constraints.
\newblock In: \emph{2019 IEEE/RSJ International Conference on Intelligent Robots and Systems (IROS)}. IEEE, pp. 2672--2678.

\bibitem[{Duan et~al.(2020)Duan, Camoriano, Ferigo, Huang, Calandriello, Rosasco and Pucci}]{duan2020learning}
Duan A, Camoriano R, Ferigo D, Huang Y, Calandriello D, Rosasco L and Pucci D (2020) Learning to avoid obstacles with minimal intervention control.
\newblock \emph{Frontiers in Robotics and AI} 7: 60.

\bibitem[{Duan et~al.(2022)Duan, Victorova, Zhao, Sun, Zheng and Navarro-Alarcon}]{duan2022ultrasound}
Duan A, Victorova M, Zhao J, Sun Y, Zheng Y and Navarro-Alarcon D (2022) Ultrasound-guided assistive robots for scoliosis assessment with optimization-based control and variable impedance.
\newblock \emph{IEEE Robotics and Automation Letters} 7(3): 8106--8113.

\bibitem[{Figueroa and Billard(2018)}]{figueroa2018physically}
Figueroa N and Billard A (2018) A physically-consistent bayesian non-parametric mixture model for dynamical system learning.
\newblock In: \emph{2nd Annual Conference on Robot Learning, CoRL 2018, Z{\"u}rich, Switzerland, 29-31 October 2018, Proceedings}, volume~87.

\bibitem[{Florence et~al.(2022)Florence, Lynch, Zeng, Ramirez, Wahid, Downs, Wong, Lee, Mordatch and Tompson}]{florence2022implicit}
Florence P, Lynch C, Zeng A, Ramirez OA, Wahid A, Downs L, Wong A, Lee J, Mordatch I and Tompson J (2022) Implicit behavioral cloning.
\newblock In: \emph{Conference on Robot Learning}. PMLR, pp. 158--168.

\bibitem[{Ghasemipour et~al.(2020)Ghasemipour, Zemel and Gu}]{ghasemipour2019divergence}
Ghasemipour SKS, Zemel R and Gu S (2020) A divergence minimization perspective on imitation learning methods.
\newblock In: \emph{Conference on Robot Learning}. PMLR, pp. 1259--1277.

\bibitem[{Ho and Ermon(2016)}]{ho2016generative}
Ho J and Ermon S (2016) Generative adversarial imitation learning.
\newblock In: \emph{Advances in neural information processing systems}. pp. 4565--4573.

\bibitem[{Huang et~al.(2019)Huang, Rozo, Silv{\'e}rio and Caldwell}]{huang2019kernelized}
Huang Y, Rozo L, Silv{\'e}rio J and Caldwell DG (2019) Kernelized movement primitives.
\newblock \emph{The International Journal of Robotics Research} 38(7): 833--852.

\bibitem[{Ijspeert et~al.(2013)Ijspeert, Nakanishi, Hoffmann, Pastor and Schaal}]{ijspeert2013dynamical}
Ijspeert AJ, Nakanishi J, Hoffmann H, Pastor P and Schaal S (2013) Dynamical movement primitives: learning attractor models for motor behaviors.
\newblock \emph{Neural computation} 25(2): 328--373.

\bibitem[{Ke et~al.(2021)Ke, Choudhury, Barnes, Sun, Lee and Srinivasa}]{ke2019imitation}
Ke L, Choudhury S, Barnes M, Sun W, Lee G and Srinivasa S (2021) Imitation learning as f-divergence minimization.
\newblock In: \emph{Algorithmic Foundations of Robotics XIV: Proceedings of the Fourteenth Workshop on the Algorithmic Foundations of Robotics 14}. Springer International Publishing, pp. 313--329.

\bibitem[{Khansari-Zadeh and Billard(2011)}]{khansari2011learning}
Khansari-Zadeh SM and Billard A (2011) Learning stable nonlinear dynamical systems with {Gaussian} mixture models.
\newblock \emph{IEEE Transactions on Robotics} 27(5): 943--957.

\bibitem[{Khoramshahi et~al.(2020)Khoramshahi, Henriks, Naef, Salehian, Kim and Billard}]{khoramshahiarm}
Khoramshahi M, Henriks G, Naef A, Salehian SSM, Kim J and Billard A (2020) Arm-hand motion-force coordination for physical interactions with non-flat surfaces using dynamical systems: Toward compliant robotic massage.
\newblock In: \emph{2019 International Conference on Robotics and Automation (ICRA)}. IEEE.

\bibitem[{Kingston et~al.(2019)Kingston, Moll and Kavraki}]{kingston2019exploring}
Kingston Z, Moll M and Kavraki LE (2019) Exploring implicit spaces for constrained sampling-based planning.
\newblock \emph{The International Journal of Robotics Research} 38(10-11): 1151--1178.

\bibitem[{Kober and Peters(2014)}]{kober2014policy}
Kober J and Peters J (2014) Policy search for motor primitives in robotics.
\newblock In: \emph{Learning Motor Skills}. Springer, pp. 83--117.

\bibitem[{Kronander and Billard(2015)}]{kronander2015passive}
Kronander K and Billard A (2015) Passive interaction control with dynamical systems.
\newblock \emph{IEEE Robotics and Automation Letters} 1(1): 106--113.

\bibitem[{Kulvicius et~al.(2011)Kulvicius, Ning, Tamosiunaite and Worg{\"o}tter}]{kulvicius2011joining}
Kulvicius T, Ning K, Tamosiunaite M and Worg{\"o}tter F (2011) Joining movement sequences: Modified dynamic movement primitives for robotics applications exemplified on handwriting.
\newblock \emph{IEEE Transactions on Robotics} 28(1): 145--157.

\bibitem[{Mroueh et~al.(2012)Mroueh, Poggio, Rosasco and Slotine}]{mroueh2012multiclass}
Mroueh Y, Poggio T, Rosasco L and Slotine JJ (2012) Multiclass learning with simplex coding.
\newblock In: \emph{Advances in Neural Information Processing Systems}. pp. 2789--2797.

\bibitem[{Osa et~al.(2018)Osa, Pajarinen, Neumann, Bagnell, Abbeel and Peters}]{osa2018algorithmic}
Osa T, Pajarinen J, Neumann G, Bagnell JA, Abbeel P and Peters J (2018) An algorithmic perspective on imitation learning.
\newblock \emph{Foundations and Trends{\textregistered} in Robotics} 7(1-2): 1--179.

\bibitem[{Paraschos et~al.(2013)Paraschos, Daniel, Peters and Neumann}]{paraschos2013probabilistic}
Paraschos A, Daniel C, Peters JR and Neumann G (2013) Probabilistic movement primitives.
\newblock In: \emph{Advances in neural information processing systems}. pp. 2616--2624.

\bibitem[{Pardo(2018)}]{pardo2018statistical}
Pardo L (2018) \emph{Statistical inference based on divergence measures}.
\newblock Chapman and Hall/CRC.

\bibitem[{Peng et~al.(2018)Peng, Abbeel, Levine and van~de Panne}]{peng2018deepmimic}
Peng XB, Abbeel P, Levine S and van~de Panne M (2018) Deepmimic: Example-guided deep reinforcement learning of physics-based character skills.
\newblock \emph{ACM Transactions on Graphics (TOG)} 37(4): 143.

\bibitem[{Peters et~al.(2010)Peters, Mulling and Altun}]{peters2010relative}
Peters J, Mulling K and Altun Y (2010) Relative entropy policy search.
\newblock In: \emph{Twenty-Fourth AAAI Conference on Artificial Intelligence}.

\bibitem[{Pomerleau(1989)}]{pomerleau1989alvinn}
Pomerleau DA (1989) Alvinn: An autonomous land vehicle in a neural network.
\newblock In: \emph{Advances in neural information processing systems}. pp. 305--313.

\bibitem[{Ratliff et~al.(2009)Ratliff, Silver and Bagnell}]{ratliff2009learning}
Ratliff ND, Silver D and Bagnell JA (2009) Learning to search: Functional gradient techniques for imitation learning.
\newblock \emph{Autonomous Robots} 27(1): 25--53.

\bibitem[{Ravichandar et~al.(2020)Ravichandar, Polydoros, Chernova and Billard}]{ravichandar2020recent}
Ravichandar H, Polydoros AS, Chernova S and Billard A (2020) Recent advances in robot learning from demonstration.
\newblock \emph{Annual Review of Control, Robotics, and Autonomous Systems} 3.

\bibitem[{Reiner et~al.(2014)Reiner, Ertel, Posenauer and Schneider}]{reiner2014lat}
Reiner B, Ertel W, Posenauer H and Schneider M (2014) {LAT}: A simple learning from demonstration method.
\newblock In: \emph{2014 IEEE/RSJ International Conference on Intelligent Robots and Systems}. IEEE, pp. 4436--4441.

\bibitem[{Ross et~al.(2011)Ross, Gordon and Bagnell}]{ross2011reduction}
Ross S, Gordon G and Bagnell D (2011) A reduction of imitation learning and structured prediction to no-regret online learning.
\newblock In: \emph{Proceedings of the fourteenth international conference on artificial intelligence and statistics}. pp. 627--635.

\bibitem[{Ruan et~al.(2023)Ruan, Poblete, Wu, Ma and Chirikjian}]{9841604}
Ruan S, Poblete KL, Wu H, Ma Q and Chirikjian GS (2023) Efficient path planning in narrow passages for robots with ellipsoidal components.
\newblock \emph{IEEE Transactions on Robotics} 39(1): 110--127.
\newblock \doi{10.1109/TRO.2022.3187818}.

\bibitem[{Rudi et~al.(2015)Rudi, Camoriano and Rosasco}]{rudi2015less}
Rudi A, Camoriano R and Rosasco L (2015) Less is more: {N}ystr{\"o}m computational regularization.
\newblock \emph{Advances in Neural Information Processing Systems} 28.

\bibitem[{Rudi et~al.(2018)Rudi, Ciliberto, Marconi and Rosasco}]{rudi2018manifold}
Rudi A, Ciliberto C, Marconi G and Rosasco L (2018) Manifold structured prediction.
\newblock In: \emph{Advances in Neural Information Processing Systems}. pp. 5610--5621.

\bibitem[{Schaal(1999)}]{schaal1999imitation}
Schaal S (1999) Is imitation learning the route to humanoid robots?
\newblock \emph{Trends in cognitive sciences} 3(6): 233--242.

\bibitem[{Schneider and Ertel(2010)}]{schneider2010robot}
Schneider M and Ertel W (2010) Robot learning by demonstration with local {Gaussian} process regression.
\newblock In: \emph{2010 IEEE/RSJ International Conference on Intelligent Robots and Systems}. IEEE, pp. 255--260.

\bibitem[{Schulman et~al.(2015)Schulman, Levine, Abbeel, Jordan and Moritz}]{schulman2015trust}
Schulman J, Levine S, Abbeel P, Jordan M and Moritz P (2015) Trust region policy optimization.
\newblock In: \emph{International conference on machine learning}. pp. 1889--1897.

\bibitem[{Simo-Serra et~al.(2017)Simo-Serra, Torras and Moreno-Noguer}]{simo20173d}
Simo-Serra E, Torras C and Moreno-Noguer F (2017) 3{D} human pose tracking priors using geodesic mixture models.
\newblock \emph{International Journal of Computer Vision} 122(2): 388--408.

\bibitem[{Stulp and Sigaud(2015)}]{stulp2015many}
Stulp F and Sigaud O (2015) Many regression algorithms, one unified model: A review.
\newblock \emph{Neural Networks} 69: 60--79.

\bibitem[{Swamy et~al.(2021)Swamy, Choudhury, Bagnell and Wu}]{swamy2021moments}
Swamy G, Choudhury S, Bagnell JA and Wu S (2021) Of moments and matching: A game-theoretic framework for closing the imitation gap.
\newblock In: \emph{International Conference on Machine Learning}. PMLR, pp. 10022--10032.

\bibitem[{Traversaro et~al.(2016)Traversaro, Brossette, Escande and Nori}]{traversaro2016identification}
Traversaro S, Brossette S, Escande A and Nori F (2016) Identification of fully physical consistent inertial parameters using optimization on manifolds.
\newblock In: \emph{2016 IEEE/RSJ International Conference on Intelligent Robots and Systems (IROS)}. IEEE, pp. 5446--5451.

\bibitem[{Yang et~al.(2018)Yang, Zeng, Fang, He and Li}]{yang2018dmps}
Yang C, Zeng C, Fang C, He W and Li Z (2018) A {DMP}s-based framework for robot learning and generalization of humanlike variable impedance skills.
\newblock \emph{IEEE/ASME Transactions on Mechatronics} 23(3): 1193--1203.

\bibitem[{Zahra et~al.(2022)Zahra, Tolu, Zhou, Duan and Navarro-Alarcon}]{zahra2022bio}
Zahra O, Tolu S, Zhou P, Duan A and Navarro-Alarcon D (2022) A bio-inspired mechanism for learning robot motion from mirrored human demonstrations.
\newblock \emph{Frontiers in Neurorobotics} 16.

\bibitem[{Zeestraten et~al.(2017{\natexlab{a}})Zeestraten, Havoutis, Calinon and Caldwell}]{zeestraten2017learning}
Zeestraten MJ, Havoutis I, Calinon S and Caldwell DG (2017{\natexlab{a}}) Learning task-space synergies using {Riemannian} geometry.
\newblock In: \emph{2017 IEEE/RSJ International Conference on Intelligent Robots and Systems (IROS)}. IEEE, pp. 73--78.

\bibitem[{Zeestraten et~al.(2017{\natexlab{b}})Zeestraten, Havoutis, Silv{\'e}rio, Calinon and Caldwell}]{zeestraten2017approach}
Zeestraten MJ, Havoutis I, Silv{\'e}rio J, Calinon S and Caldwell DG (2017{\natexlab{b}}) An approach for imitation learning on {Riemannian} manifolds.
\newblock \emph{IEEE Robotics and Automation Letters} 2(3): 1240--1247.

\bibitem[{Zhou and Asfour(2017)}]{zhou2017task}
Zhou Y and Asfour T (2017) Task-oriented generalization of dynamic movement primitive.
\newblock In: \emph{2017 IEEE/RSJ International Conference on Intelligent Robots and Systems (IROS)}. IEEE, pp. 3202--3209.

\bibitem[{Ziebart et~al.(2008)Ziebart, Maas, Bagnell and Dey}]{10.5555/1620270.1620297}
Ziebart BD, Maas A, Bagnell JA and Dey AK (2008) Maximum entropy inverse reinforcement learning.
\newblock In: \emph{Proceedings of the 23rd National Conference on Artificial Intelligence - Volume 3}, AAAI'08. AAAI Press.
\newblock ISBN 9781577353683, p. 1433–1438.

\end{thebibliography}
\end{document}